\pdfoutput=1

\newcommand{\Ant}[1]{\ensuremath{\textsc{Ant}_{\mathrm{#1}}}}
\newcommand{\Syn}[1]{\ensuremath{\textsc{Syn}_{\mathrm{#1}}}}
\newcommand{\PChat}[1]{\ensuremath{\textsc{PChat}_{\mathrm{#1}}}}
\documentclass[11pt]{article}

\usepackage[]{acl}

\usepackage{subcaption}
\usepackage{caption}
\usepackage{amsmath} 
\usepackage{times}
\usepackage{latexsym}
\usepackage[table]{xcolor}
\usepackage{multirow} 
\usepackage{listings}
\usepackage[T1]{fontenc}

\usepackage[utf8]{inputenc}

\usepackage{microtype}

\usepackage{inconsolata}

\usepackage{graphicx}
\usepackage{booktabs} 
\usepackage{tabularx} 
\usepackage{lipsum} 
\usepackage{xcolor}
\usepackage{amsmath}

\title{\textsc{Synthia}: Scalable Grounded Persona Generation \\ from Social Media Data}

\author{ 
 \textbf{Vahid Rahimzadeh*\textsuperscript{1,2}},
 \textbf{Erfan Moosavi Monazzah*\textsuperscript{1}}, \\
 \textbf{Mohammad Taher Pilehvar\textsuperscript{3}}, and
 \textbf{Yadollah Yaghoobzadeh\textsuperscript{1,2}}
\\
 \textsuperscript{1}Tehran Institute for Advanced Studies, Khatam University, Iran \\
 \textsuperscript{2}University of Tehran, Iran \\
 \textsuperscript{3}Cardiff University, United Kingdom 
\\
 \texttt{\{v.rahimzade, e.moosavi\_monazzah\}@teias.institute}
}

\begin{document}
\maketitle
\begin{abstract}
\noindent
Persona-driven simulations are increasingly used in computational social science, yet their validity critically depends on the fidelity of the underlying personas. Constructing virtual populations that are both authentic and scalable remains a central challenge.
We introduce \textsc{Synthia}, a persona-generation framework that grounds LLM-generated personas in real social-media posts while delegating narrative construction to language models, using publicly available data from the Bluesky platform.
Across multiple social-survey benchmarks, \textsc{Synthia} improves alignment with human opinion distributions over prior state-of-the-art approaches while relying on substantially smaller models. A multi-dimensional fairness and bias analysis shows that \textsc{Synthia} outperforms previous methods for most demographics across different dimensions. Uniquely, \textsc{Synthia} preserves interaction-graph structure among personas grounded in real social network users, enabling network-aware analysis, which we demonstrate through two homophily-focused case studies. Together, these results position \textsc{Synthia} as a practical and reliable framework for constructing scalable, high-fidelity, and equitable virtual populations.
\end{abstract}

\section{Introduction}
\label{sec:intro}
\begingroup
  \hypersetup{hidelinks}       
  \renewcommand{\thefootnote}{*} 
  \footnotetext{Equal contribution, ordered randomly}
\endgroup

Persona-driven large language models (LLMs) are increasingly adopted across a wide range of domains~\cite{persona_survey}, particularly in population-level simulation and analysis~\cite{intro_survey_2024,intro_css_2024}. In this context, personas may range from simple demographic descriptors to rich psychological profiles and detailed life narratives~\cite{li2025llmgeneratedpersonapromise, cintas2025localizingpersonarepresentationsllms}. While explicitly conditioning models on demographic attributes can inadvertently promote stereotypical inferences and amplify bias~\cite{anthis2025llmsocialsimulationspromising}, employing context-rich personas has been shown to foster more individualized variation and reduce disparities in predictive accuracy across demographic groups~\cite{2024_1000agents_stanford}.

\begin{figure}[t]
\centering
\renewcommand{\arraystretch}{1.2}
\includegraphics[width=0.9\columnwidth]{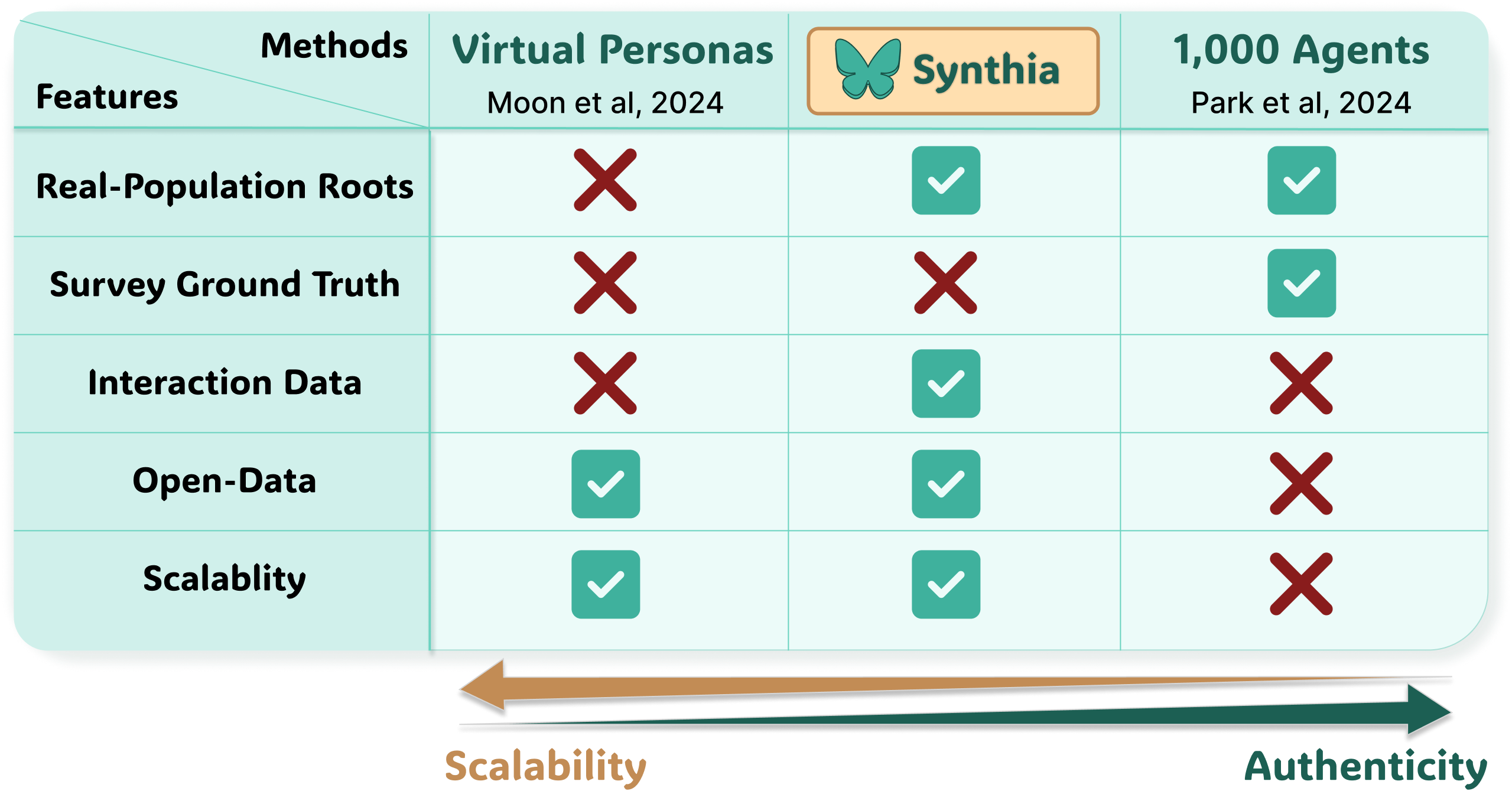}
\caption{\textsc{Synthia} vs. leading persona methods.}
\label{fig:first_page}
\end{figure}

However, scaling the construction of context-rich personas remains a central challenge, with methods falling along a spectrum between authenticity and scalability (see Figure~\ref{fig:first_page}). At one extreme are interview-based approaches that derive personas from human data and often yield improved realism~\cite{anthis2025llmsocialsimulationspromising}. Yet, these approaches are resource-intensive and difficult to scale~\cite{2024_1000agents_stanford}. In contrast, fully synthetic approaches~\cite{2024_virtual_berkeley_emnlp} offer scalability but frequently introduce systematic artifacts that reduce realism~\cite{li2025llmgeneratedpersonapromise}.

Seeking an optimal balance between scalability and authenticity, we introduce \textsc{Synthia}, \textbf{S}ynthetic \textbf{Y}et \textbf{N}aturally \textbf{T}ailored \textbf{H}uman-\textbf{I}nspired Person\textbf{A}, a methodology that grounds persona generation in real social media content. Social media contains large volumes of user-generated content reflecting diverse behaviors and viewpoints. For this work, we utilize Bluesky\footnote{\scriptsize https://bsky.app/}, due to its open platform structure and permissive redistribution policies.
Concurrently, LLMs have demonstrated a remarkable proficiency in processing such content for persona development~\cite{yin2025copersonaleveragingllmsexpert, prottasha2025userprofilelargelanguage}.

\textsc{Synthia} differs from prior work in how it uses language models. A model is used to compose a population of persona narratives from real posts, while separate models conditioned on those narratives answer demographic and opinion questions. We then match the synthetic population to the demographic composition of real survey respondents and evaluate alignment through statistical comparisons between simulated and real-world opinion distributions from social surveys. Thus, the main evaluation is anchored in human survey data rather than LLM-based judgment. An LLM judge is used only for the narrative consistency analysis, where it is validated against human annotations. This design preserves scalability while improving internal factual consistency and reducing systematic bias relative to fully synthetic pipelines, which are key factors for better alignment and fidelity to human populations. Unlike prior works that provide persona text only, \textsc{Synthia} also supplies interaction-graph metadata, enabling network-aware analyses.

Our comprehensive evaluation across 54 experimental configurations establishes \textsc{Synthia} as a robust alternative to SOTA methods. In terms of population opinion alignment, it improves upon baselines by up to 11.6\% across social surveys with models less than half the size. Error analysis suggests that these gains are largely driven by a reduction in inner narrative factual contradictions per persona. Furthermore, our fairness analysis reveals that \textsc{Synthia} consistently achieves high fidelity while maintaining stability across demographic groups, reducing accuracy gaps between best and worst performing subgroups by up to 25\%, and minimizing disparate impacts in sensitive categories like gender and education.
We further demonstrate \textsc{Synthia}'s applicability to social network analysis. By preserving the source topology, our personas effectively encode the correlation between structurally \& semantically informed homophily, achieving accuracy gains of 8.3\% ($p < 0.001$) in link prediction tasks and increasing embedding-space separability between connected and unconnected personas by up to 46\%.

Our contributions are fourfold: (i) We propose \textsc{Synthia}, a scalable persona-generation pipeline that produces representative, human-like virtual populations grounded in real social-media content. (ii) We show that internal factual consistency is critical for accurately modeling population-level opinions and diversity. (iii) We demonstrate that grounding reduces systematic bias and improves fairness, enabling reliable persona generation with substantially smaller language models. (iv) We release a large-scale dataset of grounded virtual personas together with their underlying social interaction graph, and illustrate its utility through downstream computational social science case studies.

\section{Related Work}
\label{sec:related}
Persona-driven use cases of LLMs have been the focus of numerous recent studies~\cite{anthis2025llmsocialsimulationspromising, intro_survey_2024, intro_bias_2024, intro_css_2024}, covering aspects such as the strengths and biases of LLMs ~\cite{suh-etal-2025-language,2024_incongrnous_steered_cmu,chen-etal-2024-socialbench,2023_icl_explore_exploit_nips,2023_compost_stanford_emnlp,2023_opinionqa_icml,Argyle_2023}, computational social science simulations ~\cite{piao2025agentsocietylargescalesimulationllmdriven,shen-etal-2025-words,2024_social_simulation_cocoons_conformity_tois,2024_social_manipulation_rabbany,rahimzadeh2025millions,Argyle_2023}, policy and governance decision-making ~\cite{2024_cooperate_policy_llm_nips, 2024_policy_impact_simulation}, and user behavior modeling ~\cite{suh2025rethinkingllmhumansimulation,he2025simupanelnovelimmersivemultiagent,wang-etal-2025-know,2024_community_persona_emnlp,2023_simulacra_stanford}.
As the applications of persona-driven models expand, more research has emerged on methodologies for creating these personas~\cite{10.1145/3706598.3713445, bui-etal-2025-mixture, bückkaeffer2025textttblueprintsocialmediauser, liu-etal-2025-mosaic, JUNG2025103445, yin2025copersonaleveragingllmsexpert, dash2025polypersonapersonagroundedllmsynthetic}. Despite current attempts at creating personas through role playing~\cite{intro_survey_2024, intro_bias_2024, intro_css_2024}, in-context learning~\cite{2024_elicit_persona_bayes_icml, 2023_icl_explore_exploit_nips} or aligning models to specific sets of opinions from real users~\cite{2023_align_opinion_vertex_emnlp, 2023_opinionqa_icml}, the existing approaches still leave an important gap in scalable methods that combine rich narrative structure with grounding in real user-generated evidence.

One approach is to condition LLMs on backstories that encode a life narrative~\cite{2024_1000agents_stanford, 2024_virtual_berkeley_emnlp}. Life narratives provide a structured representation of identity, reflecting demographic and social attributes such as gender, ethnicity, and social class~\cite{2024_virtual_berkeley_emnlp, narr_2024, narr_2013}.
Recent work has emphasized grounding these narratives in real human data to improve authenticity, e.g., \citet{2024_1000agents_stanford} simulate the attitudes and behaviors of real individuals by applying LLMs to qualitative interview data and evaluating how well the resulting agents reproduce observed human responses.

\begin{figure*}[!t]
\centering
\renewcommand{\arraystretch}{1.2}
\includegraphics[width=1\textwidth]{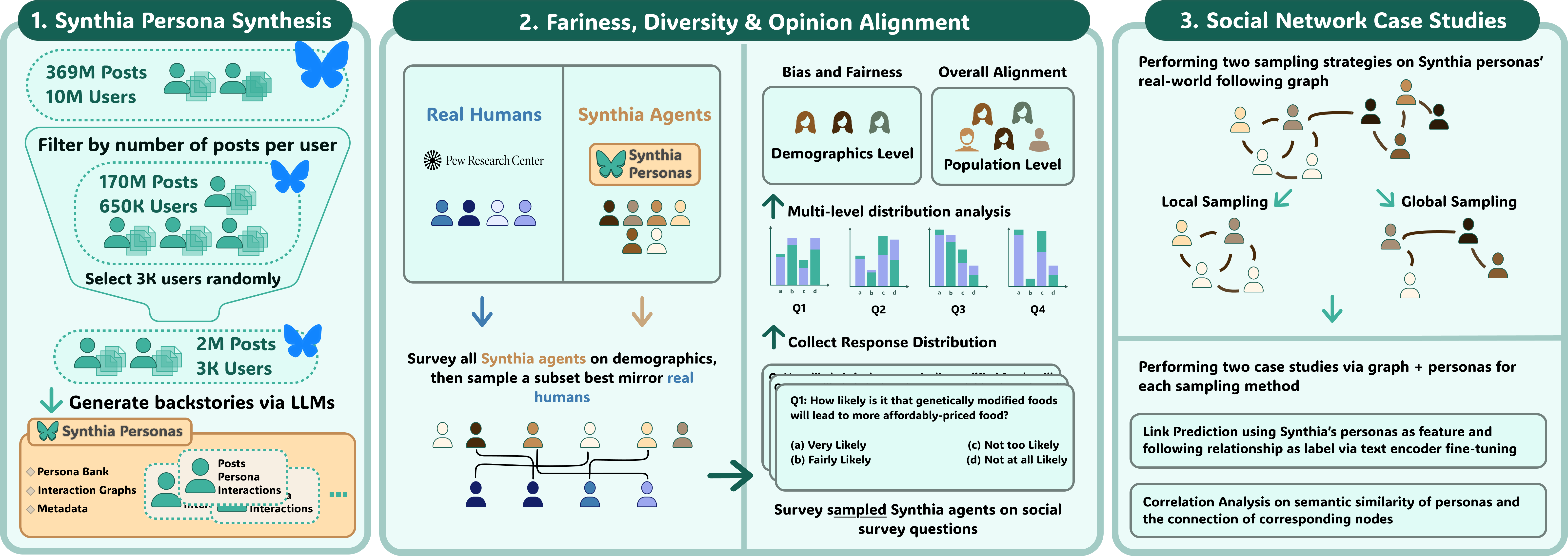}
\caption{Our approach involves: 1) Collecting and filtering high-quality user data from an open social network and generating \textsc{Synthia} Personas (Sec 3), 2) Evaluating population diversity \& opinion alignment with real-world social surveys (Sec 4) and Bias Analysis on the performance and stability across demographics (Sec 5) and 3) Case studies on homophily with social networks (Sec 6).}
\label{fig:method}
\end{figure*}

\citet{2024_virtual_berkeley_emnlp} use high-temperature LLM sampling to generate diverse life narratives, exploiting models' broad distributional knowledge. However, unconstrained generation without real-world grounding risks hallucination and narrative inconsistency.

\section{Synthia Persona Synthesis}
\label{sec:synthesis}
In this work, we define a persona as a concise first-person life narrative that articulates the backstory of a virtual individual, including preferences, salient life events, and other relevant biographical details. This definition is consistent with prior work on both synthetic~\cite{2024_virtual_berkeley_emnlp} and human-authored personas~\cite{persona_chat}, enabling direct comparison with existing approaches. Figure~\ref{fig:method} gives an overview of our methodology.
\paragraph{User Pool Creation.}
To ground \textsc{Synthia} personas in real-world data, we curate a diverse pool of social-media content consisting of posts and users from the open platform Bluesky, selected due to its permissive licensing terms that allow public redistribution (see Appendix~\ref{appendix:dataset_generation}). All selected data are normalized into a unified schema, including de-duplication and the removal of non-English content. We further filter users with atypical posting behavior by excluding accounts whose activity is either too sparse or too dense. Through preliminary analysis, we observe that users with fewer than 100 posts over a two-year period provide insufficient context, often resulting in overly brief personas or forcing the LLM to hallucinate content. Conversely, users with more than 1{,}000 posts exceed the context window of the persona generator model (Appendix~\ref{appendix:dataset_generation}). After filtering, our dataset contains approximately 170M posts from 650K unique users. 
To ensure fair comparison with prior work~\cite{2024_virtual_berkeley_emnlp,persona_chat}, we take a random sample of users from this pool with about the same size as the smallest baseline, which is 3K.

\paragraph{Persona Generation.}
To demonstrate the scalability of our approach, we employ a lightweight open LLM that can be deployed on consumer-grade GPUs. Specifically, we use Gemma-3-27B~\cite{gemma_3}. To further assess scalability and analyze the effect of model parameter count on persona backstory generation, we also include a smaller language model (SLM), Phi-4-mini~\cite{phi_4_mini} (see Appendix~\ref{appendix:experimental_setups} for full experimental details). Before persona generation, we remove explicit social identifiers and interaction cues from the source text, including handles, mentions, URLs, emails, phone numbers. We also exclude replies/reposts from the generation corpus used for persona construction.
Using the collected social-media posts for each user, we prompt these models (Figure~\ref{fig:pt_gen}) to generate comprehensive first-person backstories. An example illustrating the grounding between source posts and the resulting persona is shown in Figure~\ref{fig:grounding}. To isolate the effect of model size in comparative evaluations, we re-run the persona generation pipeline from the current state-of-the-art synthetic persona framework~\cite{2024_virtual_berkeley_emnlp} using Gemma-3-27B. While the original backstories in that work were generated using Llama-3-70B, employing the same 27B model across both pipelines ensures a controlled and fair comparison. Dataset statistics for all persona collections are reported in Table~\ref{tab:combined_dataset_stats}. 


\paragraph{Social Network Graph.}
A unique characteristic of \textsc{Synthia} is the underlying interaction graph among generated personas. This structure is directly inherited from the original social network of the users selected for synthesis. We formally represent the network as a directed graph, where a directed edge denotes a following relationship between users. This representation enables analyses that jointly consider persona and network structure.

\section{Diversity \& Opinion Alignment}
\label{sec:alignment}
A primary use of virtual personas is their ability to respond to social surveys as proxies for human respondents. This capability enables the evaluation of fidelity, that is, how faithfully a synthetic population reflects human distributions of social attitudes and behaviors. To assess this alignment, we evaluate personas using surveys from the American Trends Panel (ATP). For direct comparability with prior work, we focus on Wave~34 and Wave~99 of the ATP datasets~\cite{pew_atp_w34,pew_atp_w99}. After matching personas to the demographic composition of survey respondents, we compare the simulated opinion distributions of matched personas to real-world survey distributions using statistical metrics. Throughout this section, we use non-instruction-tuned models, as prior work has shown them to outperform instruction-tuned variants in survey-response simulation~\cite{2024_virtual_berkeley_emnlp}. Details regarding survey questions and experimental configurations are provided in Appendix~\ref{appendix:wave_qs}.

\subsection{Demographic Matching}
To accurately simulate social surveys, the persona population must reflect the demographic composition of the survey respondents. We adopt the demographic matching procedure of \citet{2024_virtual_berkeley_emnlp}, selecting a subset of personas whose aggregate demographic distribution aligns with that of the target population.

For each persona, we infer demographic attributes through \emph{demographic surveying}, in which an LLM is conditioned on the persona’s backstory and repeatedly queried with demographic questions. This yields a stable probability distribution over demographic attributes per persona. We then use a greedy matching algorithm to assign each survey respondent to the persona whose inferred demographic profile most closely matches their own. Details of the matching procedure are provided in Appendix~\ref{appendix:atp_details}. 

\subsection{Opinion Alignment Evaluation}
After demographic matching, we perform opinion surveying on the matched synthetic population. For each persona, we prompt the LLM to generate responses to the wave-specific opinion questions (see Appendix~\ref{appendix:wave_qs}). To quantify alignment between synthetic and real populations, we compare the resulting opinion distributions using the standard metrics introduced by~\citet{2024_virtual_berkeley_emnlp}: Earth Mover’s Distance (EMD), Frobenius Norm (Frob.), and Cronbach’s Alpha (Cron.). Thus, the main alignment results are obtained by statistical comparison to empirical human response distributions, not by LLM-based judges.

\subsection{Experiments and Results}
We conduct demographic matching and opinion surveying for five persona sets: (1) \Syn{Gemma}, our primary \textsc{Synthia} personas; (2) \Syn{Phi}, \textsc{Synthia} personas generated using a smaller model; (3) \PChat{Human}, human-authored personas~\cite{persona_chat}; (4) \Ant{Gemma}, Anthology personas~\cite{2024_virtual_berkeley_emnlp} generated using the same model as \Syn{Gemma}; and (5) \Ant{LLaMa}, the original Anthology personas.

\begin{table}[t!]
\centering
\small
\setlength{\tabcolsep}{5pt}
\begin{tabular}{llccc}
\toprule
\textbf{Wave} & \textbf{Exp.} & \textbf{EMD} $\downarrow$ & \textbf{Frob.} $\downarrow$ & \textbf{Cron. $\alpha$} $\uparrow$ \\
\midrule
\multirow{6}{*}{\textbf{W34}} 
& \Ant{LLaMa} & \textbf{0.34} & \underline{2.41} & \underline{0.35} \\
& \Ant{Gemma} & \underline{0.34} & 2.65 & 0.32 \\
& \Syn{Gemma} & 0.36 & \textbf{2.25} & \textbf{0.38} \\
& \Syn{Phi} & 0.38 & 2.61 & 0.31 \\
& \PChat{Human} & 0.35 & 2.76 & 0.29 \\ 
\addlinespace[1pt] \cmidrule{2-5} \addlinespace[1pt] 
& \textit{Human} & \textit{0.06} & \textit{0.42} & \textit{0.78} \\
\midrule
\multirow{6}{*}{\textbf{W99}} 
& \Ant{LLaMa} & \underline{0.37} & \textbf{2.03} & \textbf{0.41} \\
& \Ant{Gemma} & 0.49 & 2.41 & 0.20 \\
& \Syn{Gemma} & \textbf{0.34} & 2.21 & 0.34 \\
& \Syn{Phi} & 0.45 & 2.39 & 0.17 \\
& \PChat{Human} & 0.58 & \underline{2.05} & \underline{0.38} \\ 
\addlinespace[1pt] \cmidrule{2-5} \addlinespace[1pt] 
& \textit{Human} & \textit{0.08} & \textit{0.33} & \textit{0.83} \\
\bottomrule
\end{tabular}
\caption{Screening stage results per wave. Best values per wave are \textbf{bolded}, second-best are \underline{underlined}.}
\label{tab:screening_results}
\end{table}

\paragraph{Screening Stage.}
Given the high computational cost of full-scale evaluations, we conduct this preliminary screening to filter persona generation methods before proceeding to detailed analysis. To isolate the quality of the personas, we utilize a fixed LLM for response generation across all conditions (see Appendix~\ref{appendix:opinion_alignment_res}). Table~\ref{tab:screening_results} presents the results.

We first compare our method against our primary competitor of comparable size, \textbf{\Ant{Gemma}}. \textbf{\Syn{Gemma}} demonstrates decisive superiority, consistently outperforming \Ant{Gemma} in \textit{Frob.} and  \textit{Cron.}, which indicates stronger structural alignment and internal consistency. In terms of  \textit{EMD}, while \Syn{Gemma} maintains parity in Wave 34, it significantly surpasses the competitor in Wave 99 ($0.34$ vs. $0.49$). Remarkably, even our smallest model, \textbf{\Syn{Phi}}, proves comparable to the \Ant{Gemma} baseline across waves, despite being generated by a model approximately six times smaller (4B vs. 27B).

Next, we examine the human-generated \textbf{\PChat{Human}} baseline. While \PChat{Human} performs well in Wave 99, it proves highly volatile compared to the stability of our method. \PChat{Human} exhibits drastic cross-wave fluctuations (e.g., $\Delta_{\text{Frob.}} = 0.71$) compared to the negligible variance of \Syn{Gemma} ($\Delta_{\text{Frob.}} = 0.04$), suggesting that the human-generated baseline suffers from representational inconsistencies.

Finally, \textbf{\Syn{Gemma}} performs neck-and-neck with the significantly larger state-of-the-art model, \textbf{\Ant{LLaMa}}. Despite \Ant{LLaMa} utilizing a generator over twice the size (70B vs. 27B), the leadership is perfectly split: out of the six best performance scores recorded across waves and metrics ($3 \text{ metrics} \times 2 \text{ waves}$), three are secured by \Syn{Gemma} and three by \Ant{LLaMa}. Consequently, we retain \textbf{\Ant{LLaMa}} and \textbf{\Syn{Gemma}} as primary candidates, alongside \textbf{\Syn{Phi}} to analyze the impact of generator scale, allowing us to rigorously evaluate our proposed approach against SOTA methods.

\paragraph{Detailed Analysis.}
To robustly evaluate the quality of persona sets, we selected three distinct LLMs to serve as both demographic surveyors and response surveyors (setup details in Appendix~\ref{appendix:experimental_setups}). This factorial design, spanning three models for both survey roles, across two waves and three persona sets, yielded a total of 54 experiments ($3 \times 3 \times 2 \times 3$), or 27 per wave. We computed the mean performance per persona set per wave, with a comprehensive overview provided in Table~\ref{tab:rigorous_results}.

\begin{table}[t!]
\centering
\small
\setlength{\tabcolsep}{4pt} 
\scalebox{0.93}{
\begin{tabular}{clccc}
\toprule
 & \textbf{Exp.} & \textbf{EMD} $\downarrow$ & \textbf{Frob.} $\downarrow$ & \textbf{Cron. $\alpha$} $\uparrow$ \\
\midrule
\multirow{4}{*}{\rotatebox{90}{\textbf{W34}}} 
& \Ant{LLaMa} & \underline{0.35 $\pm$ 0.02} & 2.46 $\pm$ 0.06 & 0.34 $\pm$ 0.05 \\
& \Syn{Gemma} & ~~\textbf{0.35 $\pm$ 0.03}* & \textbf{2.30 $\pm$ 0.20} & \textbf{0.39 $\pm$ 0.09} \\
& \Syn{Phi} & 0.38 $\pm$ 0.04 & \underline{2.43 $\pm$ 0.10} & \underline{0.38 $\pm$ 0.06} \\ 
\addlinespace[1pt] \cmidrule{2-5} \addlinespace[1pt] 
& \textit{Human} & \textit{0.06} & \textit{0.42} & \textit{0.78} \\
\midrule
\multirow{4}{*}{\rotatebox{90}{\textbf{W99}}} 
& \Ant{LLaMa} & 0.43 $\pm$ 0.06 & \underline{2.14 $\pm$ 0.20} & \underline{0.35 $\pm$ 0.10} \\
& \Syn{Gemma} & \textbf{0.38 $\pm$ 0.06} & \textbf{2.12 $\pm$ 0.15} & \textbf{0.39 $\pm$ 0.09} \\
& \Syn{Phi} & \underline{0.41 $\pm$ 0.04} & 2.21 $\pm$ 0.15 & 0.31 $\pm$ 0.09 \\ 
\addlinespace[1pt] \cmidrule{2-5} \addlinespace[1pt] 
& \textit{Human} & \textit{0.08} & \textit{0.33} & \textit{0.83} \\
\bottomrule
\end{tabular}
}
\caption{Detailed analysis results per wave. *At three decimal places, \Syn{Gemma} outperforms \Ant{LLaMa}.}
\label{tab:rigorous_results}
\end{table}

Crucially, the superior performance of \textsc{Synthia} is robust to the choice of surveyor. When disaggregating results across the factorial design, \Syn{Gemma} achieves the top rank in every tested configuration, recording the lowest EMD scores regardless of which model is employed as the Demographic or Response surveyor. For instance, while \Ant{LLaMa} and \Syn{Phi} frequently exhibit EMD scores hovering around 0.40, \Syn{Gemma} maintains consistently tighter alignment, reaching a minimum EMD of 0.33.  Detailed heatmaps visualizing these surveyor-specific dynamics for EMD, Frob, and Cron and the full results across experimental settings are provided in Appendix~\ref{appendix:opinion_alignment_res}.

\paragraph{Sensitivity to Likert Scale Resolution.}
To assess whether our conclusions depend on fine-grained Likert calibration, we perform a sensitivity analysis using a coarser 3-point ordinal scale. Specifically, we collapse the original 5-point response categories into 3-point bins and recompute EMD, Frob., and Cron.\ $\alpha$ across the same evaluation settings. The qualitative ordering of methods remains largely unchanged under this coarser scale: \Syn{Gemma} retains its lead in five of the six primary wave-level comparisons, and the correspondence between the 5-point and 3-point evaluations remains high across settings (Spearman $\rho \approx 0.87$; mean Pearson $r = 0.89$). These results suggest that the gains of \Syn{Gemma} reflect robust improvements in opinion alignment rather than artifacts of fine-grained response calibration. Full results are provided in Appendix~\ref{appendix:likert_sensitivity}.

\begin{table*}[th!]
\centering
\small
\setlength{\tabcolsep}{4pt}
\scalebox{0.95}{
\begin{tabular}{@{} p{2cm} p{14cm} @{}}
\toprule
\textbf{Source} & \textbf{Narratives with  contradictions highlighted} \\
\midrule
\textbf{Anthology} & 
Growing up, I found solace in the magical worlds of Disney movies ... My love for these films began with classics like \textbf{`The Lion King' and `Mary Poppins,'} which I watched with my parents. {\textbf{These movies, released around the same time}}, shared a similar vibe... \\
\midrule
\textbf{Synthia} & 
I was \textbf{born into a wealthy family} in city X... met my wife in university studying psychology. {\textbf{My parents were immigrants so I want to help them out with living expenses}}. I dislike non-fiction books. \\
\bottomrule
\end{tabular}
}
\caption{Examples of narratives with inconsistencies. Bolded text indicates the contradicting statesments.}
\label{tab:inconsis_examples}
\end{table*}

\paragraph{Persona Consistency Analysis.}
To better understand the performance gap between \textsc{Synthia} and purely synthetic baselines, we analyze internal factual consistency within persona narratives. Manual inspection revealed that Anthology personas frequently contain internally contradictory statements (Table~\ref{tab:inconsis_examples}). To systematically quantify this issue, we apply line-to-line inconsistency detection~\cite{abdulhai2025consistentlysimulatinghumanpersonas} across all personas.

We use an LLM-based judge to identify inconsistent text spans within each persona. To ensure the reliability of this judge, we compare its outputs against human annotations on a subset of personas and benchmark three state-of-the-art API LLMs, selecting the model with the highest agreement with human judgments (69\%; see Table~\ref{tab:human_annotator_aggreement} and Appendix~\ref{appendix:consistency_analysis}). The resulting statistics are summarized in Table~\ref{tab:dataset_contradictions}.

\begin{table}[t!]
\centering
\small
\setlength{\tabcolsep}{8pt}
\begin{tabular}{l c c}
\toprule
\multirow{2}{*}{\textbf{Dataset}} & \textbf{\% Personas with} & \textbf{Mean Error} \\
& \textbf{Contradiction} & \textbf{per Persona} \\
\midrule
\Ant{LLaMa} & 0.63 & 0.959 \\
\Syn{Gemma} & 0.18 & 0.221 \\
\midrule
\PChat{Human} & 0.04 & 0.047 \\
\bottomrule
\end{tabular}
\caption{Results of persona contradiction analysis.}
\label{tab:dataset_contradictions}
\end{table}

Using human-authored personas (\PChat{Human}) as a reference, we find that \Ant{LLaMa} contains more than three times as many personas with at least one internal contradiction compared to \Syn{Gemma} (0.63\% vs.\ 0.18\%). Because individual personas may contain multiple contradictory spans, we also compute the mean number of inconsistencies per persona. Under this metric, \Syn{Gemma} again outperforms \Ant{LLaMa} by a large margin, indicating substantially improved internal factual consistency.

\paragraph{Persona Faithfulness to Underlying Posts.}
To complement our analysis of internal consistency, we evaluate the faithfulness 
of generated personas to their underlying social media posts. Following established 
methods in factuality evaluation~\cite{min-etal-2023-factscore, laban-etal-2022-summac, 
honovich-etal-2021-q2}, we formulate this task as an atomic evaluation of factual precision over generated personas.

\begin{table}[t!]
\centering
\small
\setlength{\tabcolsep}{8pt}
\begin{tabular}{l c c}
\toprule
\multirow{2}{*}{\textbf{\# Posts Used}} & \textbf{Faithfulness} & \textbf{Hallucination} \\
& \textbf{($\mathcal{F}$)} & \textbf{Rate ($\mathcal{H}$)} \\
\midrule
$< 20$       & 0.5762 & 0.4237 \\
$20$--$50$   & 0.6244 & 0.3755 \\
$50$--$100$  & 0.7078 & 0.2921 \\
$100$--$200$ & 0.7518 & 0.2481 \\
$> 200$      & 0.7534 & 0.2465 \\
\bottomrule
\end{tabular}
\caption{Persona faithfulness to underlying social media posts across five ranges of post counts used during generation.}
\label{tab:faithfulness}
\end{table}

Specifically, we (1) decompose each generated persona into a set of atomic claims 
$\mathcal{C} = \{c_1, c_2, \ldots, c_N\}$, (2) retrieve the most relevant posts from 
the user's history for each claim, (3) apply an LLM-based entailment classifier to 
label each claim as \textit{supported}, \textit{contradicted}, or \textit{unverifiable} 
given the retrieved evidence, and (4) compute faithfulness and hallucination rate, 
defined as $\mathcal{F} = S/N$ and $\mathcal{H} = (C + U)/N$, respectively, where 
$N = |\mathcal{C}|$ denotes the total number of atomic claims, $S$ the number of 
supported claims, $C$ the number of contradicted claims, and $U$ the number of 
unverifiable claims. Note that $S + C + U = N$ by construction, so 
$\mathcal{F} + \mathcal{H} = 1$.

We conduct this experiment on a sample of 250 personas across five distinct ranges of 
post counts used during generation. To ensure the reliability of the automated 
evaluation, we manually inspect the LLM outputs at each stage of the pipeline and 
correct any errors.

The results (see Table~\ref{tab:faithfulness}) reveal a clear positive correlation 
between the number of posts used to generate a persona and its faithfulness to those 
posts. This finding suggests that providing the generator with more user content 
reduces the incidence of hallucinated claims.

\section{Bias and Fairness Analysis}
\label{sec:bias}
We present a comprehensive evaluation of fairness and bias in \textsc{Synthia} relative to strong baselines. Our analysis examines subgroup-level fidelity, stability, and parity, assessing whether \textsc{Synthia} personas represent diverse demographics with comparable accuracy and reliability. We ground this evaluation in established frameworks for algorithmic fairness~\cite{bias_hardt_book}, focusing in particular on \emph{representational harms}, where models may oversimplify minority groups or fail to capture within-group nuance.

\begin{figure}[t!]
    \centering
    \includegraphics[width=0.99\linewidth]{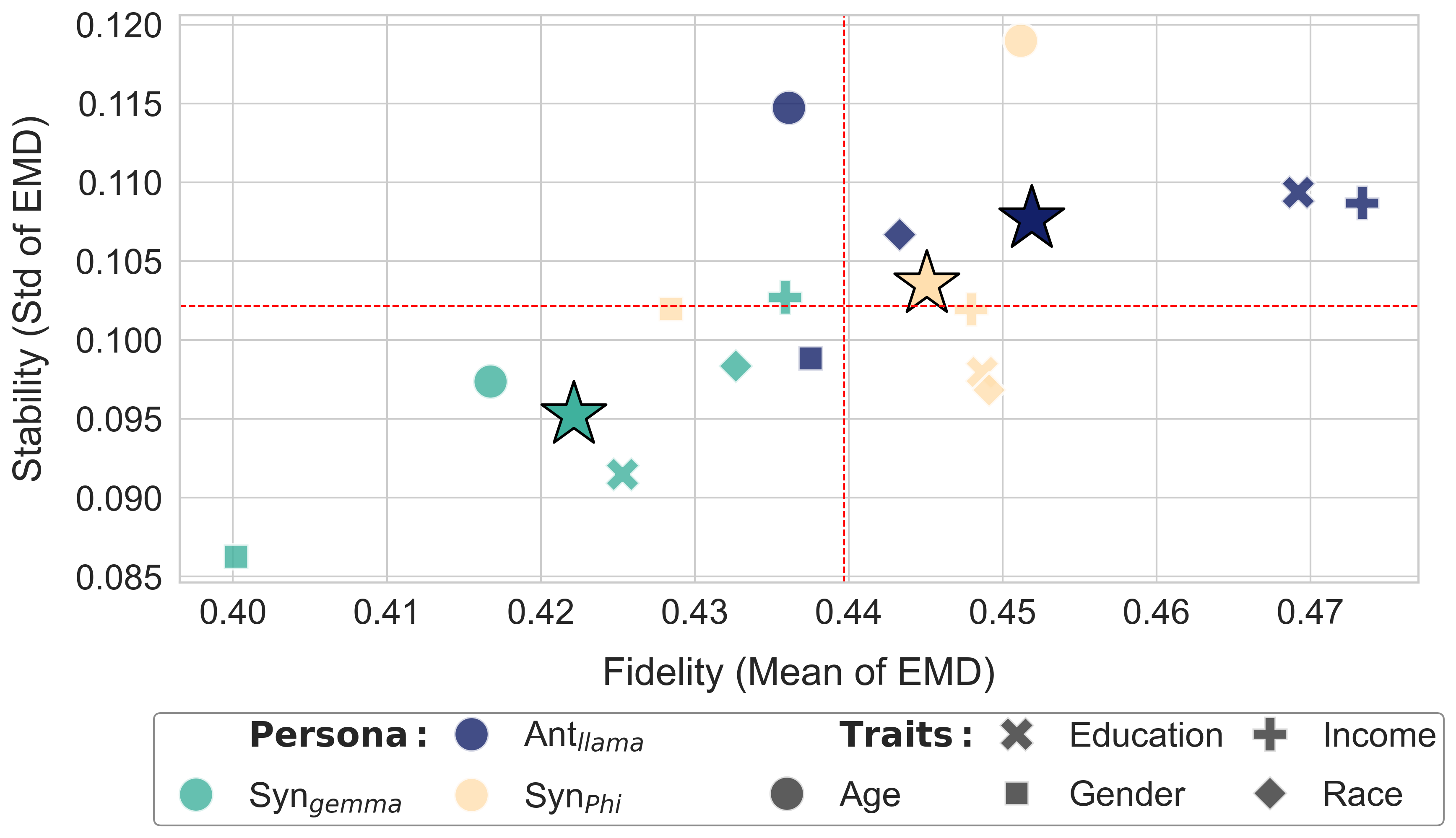}
    \caption{Relation between Fidelity and Stability (for both, lower is better). Red lines are the global average. Stars are the overall averages for each persona type.}
    \label{fig:fidelity_stability}
\end{figure}

\paragraph{Fidelity and Stability.}
We first analyze the relationship between fidelity and stability across demographic subgroups. Fidelity measures how closely synthetic personas reflect the opinions of their real-world counterparts, while stability captures the consistency of this alignment across generations. Figure~\ref{fig:fidelity_stability} visualizes this relationship. We quantify fidelity using the mean Earth Mover’s Distance (EMD) per demographic group, and stability using the standard deviation of these EMD scores.

As shown in Figure~\ref{fig:fidelity_stability}, \Syn{Gemma} consistently occupies the ``Ideal'' (bottom-left) quadrant. Averaged across all demographics, \Syn{Gemma} attains a lower mean EMD than \Ant{LLaMa}, indicating improved fidelity. In addition, \Syn{Gemma} exhibits lower variance across runs, reflecting greater stability. Together, these results show that while \Ant{LLaMa} can achieve reasonable performance in some settings, \Syn{Gemma} produces more reliable and consistently faithful representations across demographic groups.

\begin{figure}[t!]
    \centering
    \includegraphics[width=1\linewidth]{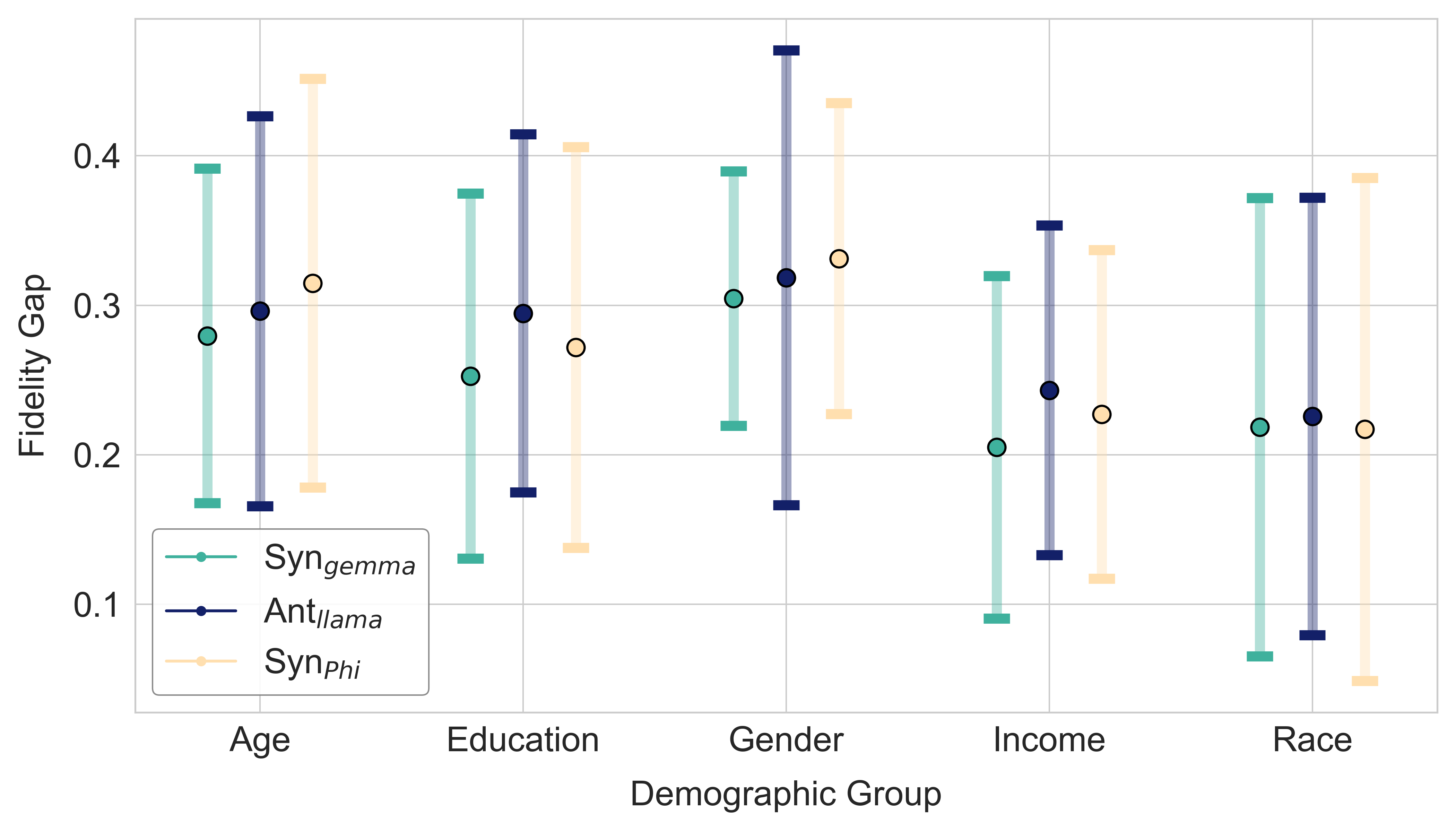}
    \caption{Subgroup Fidelity gap analysis. Values near zero indicate performance matching human variation.}
    \label{fig:fidelity_gap}
\end{figure}

\paragraph{Relative Fidelity and Human Baselines.}
Although raw EMD scores quantify distributional differences, they do not account for the inherent diversity and variance within real human populations across demographic subgroups. Groups with highly polarized opinions are intrinsically more difficult to simulate than those exhibiting broad consensus. To account for this, we normalize model fidelity against real-world human opinion distributions.
Specifically, we compute the \emph{Fidelity Gap} as the difference between a model’s EMD for a given subgroup and the internal EMD of the corresponding human subgroup, which serves as a lower bound on natural human variation (see Appendix~\ref{appendix:fidelity_gap} for implementation details).

Figure~\ref{fig:fidelity_gap} shows the resulting relative fidelity gaps. \Syn{Gemma} consistently exhibits smaller gaps across all demographics, with a lower average fidelity gap than \Ant{LLaMa}, indicating reduced divergence from natural human variation. This result demonstrates that \textsc{Synthia} does not merely minimize distributional distance, but does so while respecting the underlying opinion diversity within demographic subgroups.

\paragraph{Parity Gap Analysis.}
Finally, we assess whether any subgroup within a demographic is systematically disadvantaged (e.g., whether high- and low-income groups are simulated with comparable accuracy). Large performance disparities across subgroups are a well-known indicator of algorithmic bias~\cite{suvey_bias}. We define the \emph{Parity Gap} as the difference between the maximum and minimum error observed among subgroups within each demographic category (see Appendix~\ref{appendix:parity_gap} for details). Figure~\ref{fig:parity_gap} reports the Parity Gap across demographics.

\begin{figure}[t!]
    \centering
    \includegraphics[width=0.9\linewidth]{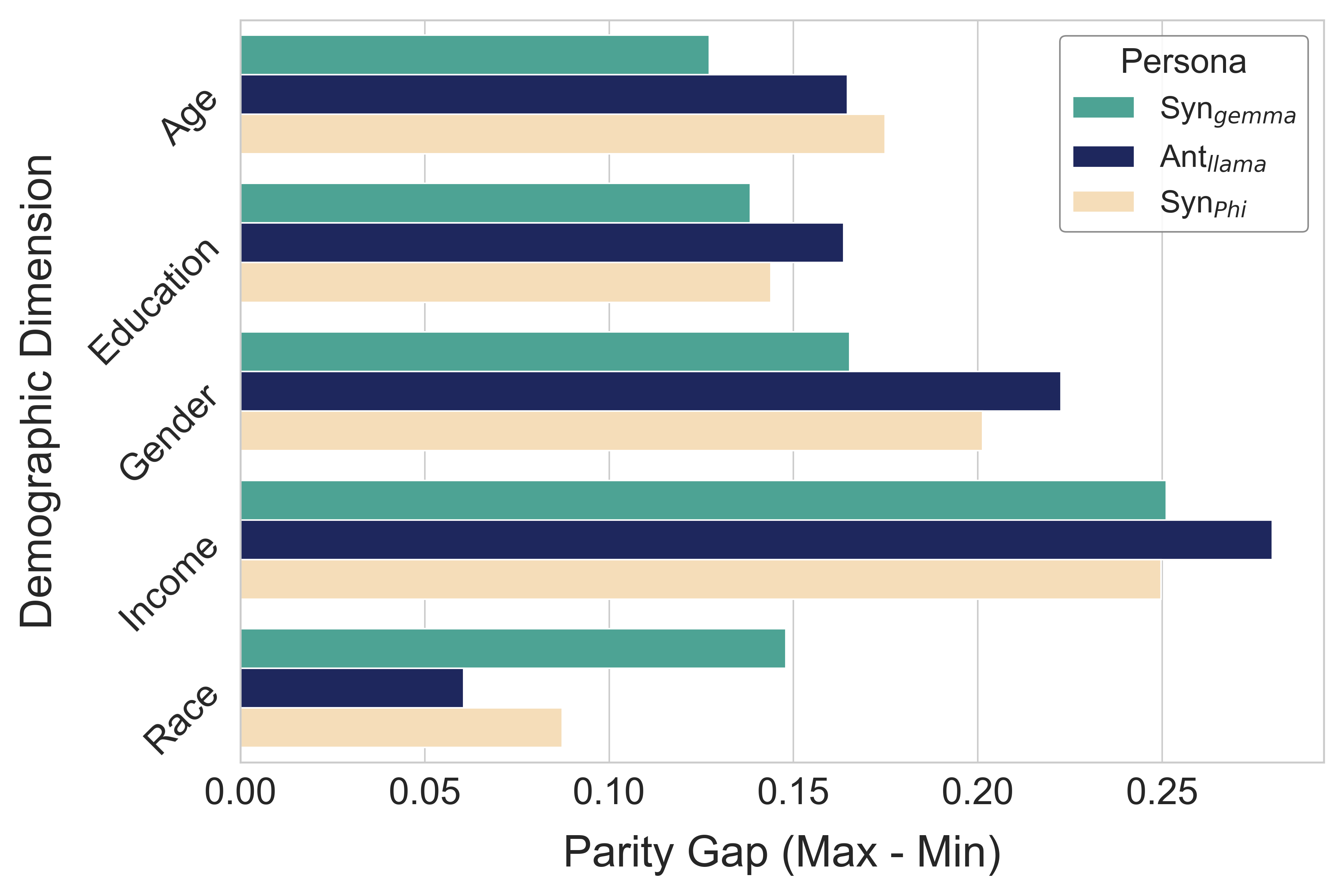}
    \caption{Parity gap analysis. Bars show the difference between best- and worst-simulated subgroups per demographic. Shorter bars indicate fairer treatment.}
    \label{fig:parity_gap}
\end{figure}

Our analysis shows that \Syn{Gemma} achieves lower parity gaps in four out of five demographic categories. For example, whereas \Ant{LLaMa} exhibits substantial disparities in Gender ($P_{\text{gap}} = 0.22$) and Income ($P_{\text{gap}} = 0.28$), \Syn{Gemma} reduces these gaps to 0.17 and 0.25, corresponding to reductions of 22.7\% and 10.7\%, respectively. These results indicate a more equitable representation of demographic subgroups and highlight the importance of reducing disparate impact when deploying synthetic personas for social science research, where representational harms must be carefully controlled.

\begin{figure}[t!]
    \centering
    \includegraphics[width=1\linewidth]{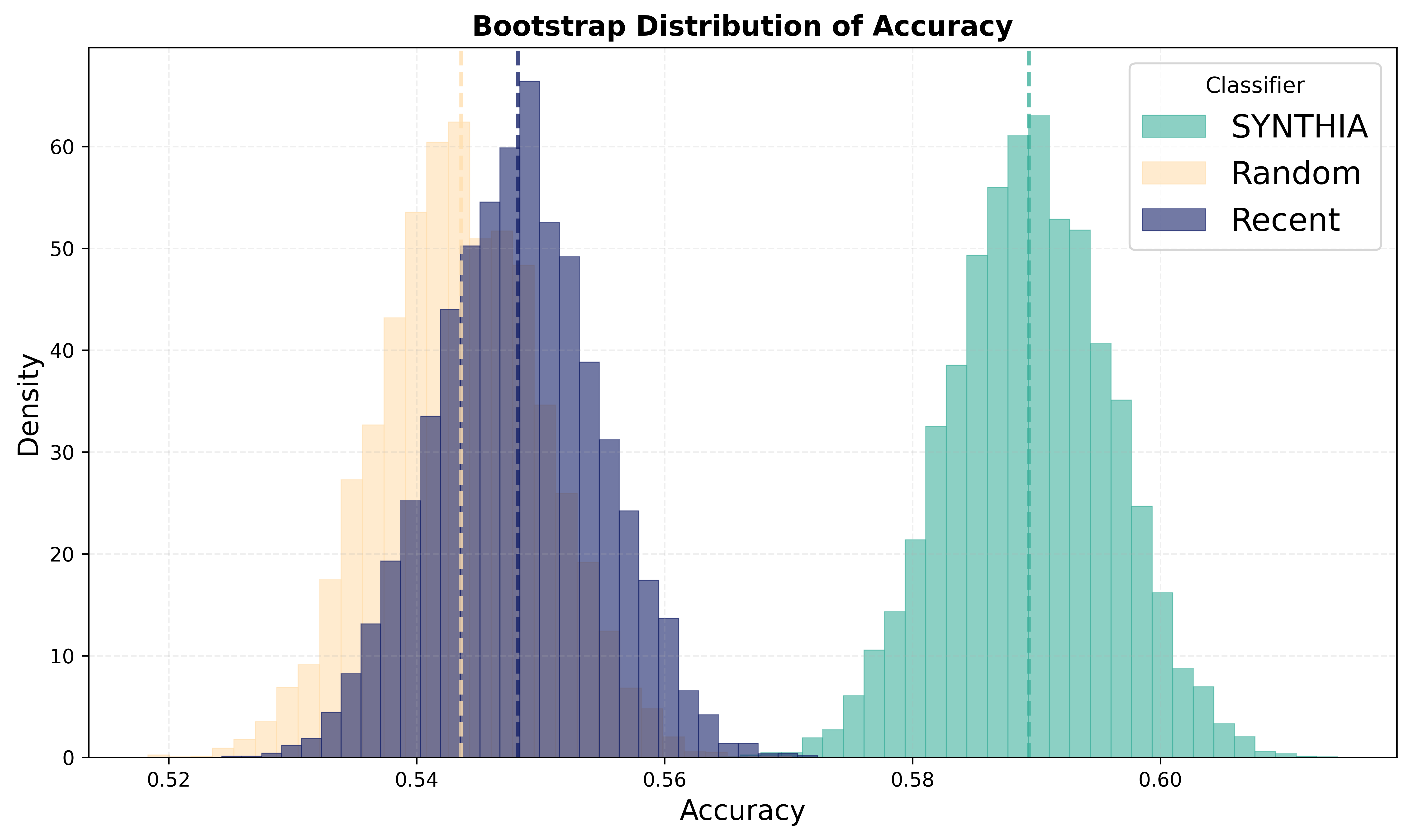}
    \caption{Bootstrap analysis of accuracy over $G_{\text{local}}$ with different baselines (N = 10,000).}
    \label{fig:boot_acc_sub}
\end{figure}

\section{Social Network Analysis}
\label{sec:cases}
To evaluate the utility of \textsc{Synthia} beyond survey simulation, we conduct two case studies on a real-world social network. We examine whether semantic similarity between personas correlates with network homophily among their corresponding nodes, assessing whether persona representations capture latent social signals aligned with observed follower–following relationships.

To capture both global and local structure, we extract two subgraphs from the full follower graph. The induced subgraph $G_{\text{global}}$, obtained via random node sampling, reflects global connectivity patterns, while the snowball-sampled subgraph $G_{\text{local}}$ captures dense, community-level interactions. Sampling details are provided in Appendix~\ref{appendix:sna_details}.

As this constitutes a new evaluation setting for \textsc{Synthia}, we construct two length-matched extractive baselines from users’ historical activity: a random baseline and a recency-based baseline that prioritizes recent posts.

\begin{table}[t]
\centering
\footnotesize 
\setlength{\tabcolsep}{10pt}
\begin{tabular}{lcccc}
\toprule
\textbf{Model} & \textbf{Acc} & \textbf{Prec} & \textbf{Rec} & \textbf{F1} \\
\midrule
\multicolumn{5}{l}{\textit{Panel A: Induced Random Subgraph ($G_{\text{global}}$)}} \\
\midrule
\textsc{Synthia} & \textbf{0.72} & \textbf{0.72} & 0.70 & \textbf{0.71} \\
Random Ext. & 0.69 & 0.68 & 0.73 & \textbf{0.71} \\
Recency Ext. & 0.68 & 0.66 & \textbf{0.74} & 0.70 \\
\midrule
\multicolumn{5}{l}{\textit{Panel B: Snowball Sampled Subgraph ($G_{\text{local}}$)}} \\
\midrule
\textsc{Synthia} & \textbf{0.59} & \textbf{0.56} & 0.91 & \textbf{0.69} \\
Random Ext. & 0.54 & 0.53 & 0.93 & 0.67 \\
Recency Ext. & 0.55 & 0.53 & \textbf{0.95} & 0.68 \\
\bottomrule
\end{tabular}
\caption{Link prediction performance on $G_{\text{global}}$ and $G_{\text{local}}$. Best results in bold; Ext: extractive.}
\label{tab:link_prediction_results}
\end{table}

\paragraph{Link Prediction.}
Our first case study evaluates whether persona text alone can predict real-world social connections. We fine-tune a transformer-based binary classifier to estimate the probability of an edge between two nodes using only the textual content of their associated personas (see Appendix~\ref{appendix:sna_details}). By excluding all explicit graph features, this setup isolates the extent to which personas capture semantically grounded homophily aligned with network structure. This evaluation is designed to test whether generated personas encode latent social similarity rather than explicit graph cues: personas are generated from privacy-filtered source posts with direct identifiers and interaction markers removed.

Table~\ref{tab:link_prediction_results} reports performance on both subgraphs. On $G_{\text{global}}$, \textsc{Synthia} outperforms both extractive baselines in accuracy and F1 score ($p < 0.001$, McNemar’s test; Panel~A), indicating that its personas encode global patterns of social similarity more effectively than activity-based summaries. Performance on $G_{\text{local}}$ is lower for all methods due to the dense, highly homophilous structure of the subgraph. The extractive baselines exhibit very high recall ($>0.93$) but low precision ($\approx 0.52$), reflecting a degenerate strategy that largely defaults to predicting positive edges.
This behavior indicates an inability to capture the fine-grained distinctions that separate actual follower relationships from general community membership.

In contrast, \textsc{Synthia} demonstrates stronger discriminative capacity, achieving higher precision and F1 score while maintaining competitive recall. As shown in Panel~B of Table~\ref{tab:link_prediction_results}, \textsc{Synthia} again outperforms both baselines ($p < 0.001$), with a larger performance gap than in $G_{\text{global}}$ (Figure~\ref{fig:boot_acc_sub}). This suggests that \textsc{Synthia} personas encode latent user interests and relational signals necessary to model fine-grained social structure.

These findings underscore that while extractive baselines prioritize coverage, evidenced by their near-universal recall on $G_{\text{local}}$, they fail to reliably distinguish true social ties from general community membership. By achieving superior precision compared to baselines, \textsc{Synthia} effectively avoids this degenerate prediction behavior. The resulting F1 score highlights the model's robustness in dense settings where determining edge existence requires capturing subtle semantic affinities rather than simple activity recency. Consequently, the statistically significant performance gains ($p < 0.001$) across both subgraphs confirm that our generated personas provide a more faithful representation of the underlying network homophily.

\begin{figure}[t!]
    \centering
    \includegraphics[width=1\linewidth]{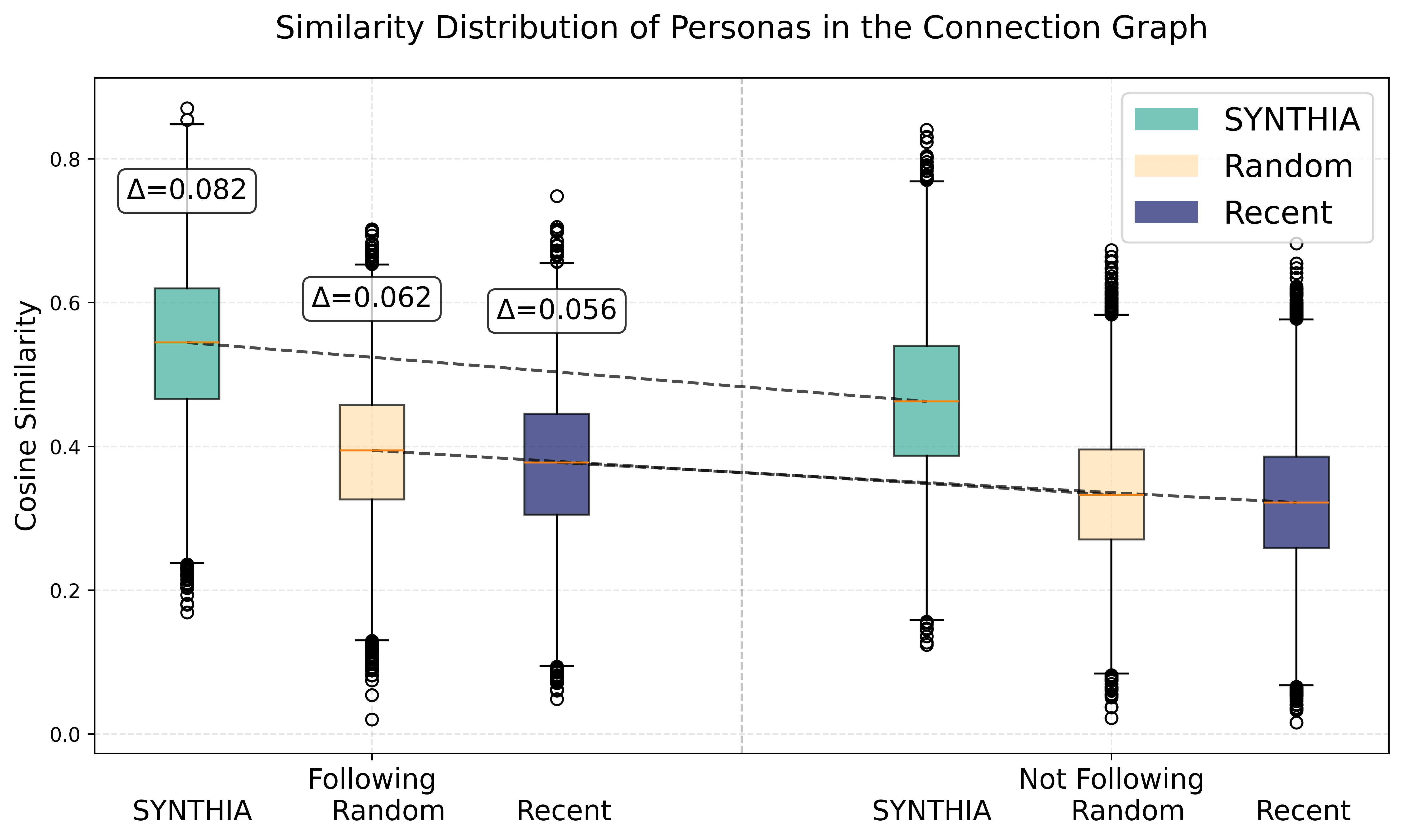}
    \caption{Semantic similarity gap between \textit{following} and \textit{not following} personas.}
    \label{fig:sim_dist_full}
\end{figure}

\paragraph{Similarity Distribution Analysis.}
To further examine the relationship between persona representations and network structure, we compute cosine similarity between persona embeddings for all node pairs with a follower relationship and an equally sized random sample of non-following pairs. This analysis is conducted for \textsc{Synthia} and both extractive baselines.
Figure~\ref{fig:sim_dist_full} shows the resulting similarity distributions. 

Across all methods, following pairs exhibit higher median similarity than non-following pairs, reflecting underlying social homophily. However, \textsc{Synthia} demonstrates substantially stronger alignment with network structure. Its personas are more semantically cohesive, with a median cosine similarity of approximately 0.55 for following pairs, compared to 0.40 and 0.38 for the random and recency baselines, respectively. In addition, \textsc{Synthia} shows the largest separation between following and non-following distributions, with a median difference of $\Delta = 0.082$, exceeding the random baseline by 32.2\% ($\Delta = 0.062$) and the recency baseline by 46.4\% ($\Delta = 0.056$).

These results indicate that \textsc{Synthia} more effectively encodes the semantic signal of social connectivity into persona representations, aligning latent textual similarity with observed network homophily. This pattern is consistent across both global and local subgraphs, suggesting it is not driven by a particular sampling strategy. While this analysis does not assume a causal relationship between similarity and edge formation, it complements the link prediction results by providing a distributional view of how persona semantics correspond to social structure.

\section{Conclusion}
We introduced \textsc{Synthia}, a persona synthesis framework that combines the scalability of large language models with grounding in real-world, human-generated data. By constructing personas from open social media content, \textsc{Synthia} addresses key limitations of purely synthetic approaches, including internal narrative inconsistency and demographic bias.

Across extensive experimental settings, we show that \textsc{Synthia} produces virtual populations that more faithfully align with population-level opinion distributions while exhibiting improved fairness and stability across sensitive demographic attributes. Moreover, by preserving the underlying social network structure of the source data, \textsc{Synthia} enables the analysis of personas within realistic relational contexts, as demonstrated through our network-based case studies.

Looking ahead, this framework opens several avenues for future research. Virtual personas grounded in social networks can be used to study interventions aimed at reducing polarization or harmful content, as well as to model diverse interaction types while accounting for network structure and temporal dynamics. We emphasize that the generated personas reflect patterns present in the underlying data and should be interpreted accordingly. To support reproducibility and further empirical investigation, we release the personas together with their associated network metadata. We hope this work encourages more principled, transparent, and responsible use of persona-driven simulations in computational social science.

\section*{Limitations}
\textit{\textsc{Synthia} should be interpreted as a tool for population and subgroup level simulation, not as a faithful reconstruction of any specific individual.}
Parts of our evaluation pipeline still rely on LLM-based simulation and LLM-based judgment. Although our primary alignment metrics are anchored to human ATP distributions rather than judge scores, survey simulation may still inherit model-specific priors, prompt sensitivities, and calibration errors. Likewise, the narrative consistency analysis uses an LLM judge with imperfect agreement to human annotators. We mitigate these risks through a factorial robustness design across surveyor models, randomized response order, use of base models for survey simulation, judge validation against human labels, and the additional faithfulness analysis introduced in this work. Still, these controls do not fully substitute for broader human-in-the-loop validation, especially for individual-level fidelity.
While \textsc{Synthia} establishes a robust framework for scalable and authentic persona generation, our current study highlights several avenues for future exploration. First, to ensure the reproducibility and accessibility of our pipeline for the broader academic community, we prioritized evaluations using open-weights models and consumer-grade hardware. While these models demonstrate high fidelity, extending the \textsc{Synthia} framework to proprietary, frontier-class models could further enhance narrative nuance, though this remains outside the scope of the current open-science focus.
Second, our validation of opinion alignment utilized the Bluesky social network due to its transparent data policies and open architecture. While this provides a rich and ethically sourced testbed, early-adopter communities on any single platform may exhibit specific sociodemographic distributions. We designed \textsc{Synthia} to be platform-agnostic; thus, applying our methodology to alternative data sources in future work could capture an even broader spectrum of global demographic variances.
Finally, our network analysis focused on static structural and semantic homophily to validate the integrity of the generated persona graph. Future research could dynamically model how these virtual personas evolve over time, offering deeper insights into temporal opinion shifts and longitudinal social dynamics.

\section*{Ethical Considerations}
The development of high-fidelity virtual personas necessitates a rigorous commitment to data privacy and responsible AI usage, and we adhered to a strict ethical framework throughout the data lifecycle. All data was sourced exclusively from the Bluesky authenticated open data stream, strictly complying with the platform's terms of service and user redistribution policies. We processed only public-facing content, respecting the ``right to be forgotten'' by excluding protected or deleted accounts.
To protect user anonymity, we implemented a multi-stage privacy pipeline prior to model ingestion. This involved utilizing automated Named Entity Recognition (NER) combined with heuristic filtering to detect and scrub Personally Identifiable Information (PII), including real names, physical addresses, and contact details from the source text. Furthermore, all original user identifiers were replaced with randomized, hashed tokens to ensure the released dataset contains no direct linkage to live social media profiles. We also obfuscated precise timestamps to prevent identification via temporal correlation attacks, retaining only the relative sequential order required to maintain narrative coherence.
Finally, we acknowledge that generative agents are dual-use technologies. While \textsc{Synthia} is designed for computational social science simulation, we explicitly prohibit its use for deceptive practices or manipulation. To this end, the released models and datasets are governed by a research-only license, and we have established a protocol to promptly remove any data points if future privacy concerns arise.

\section*{Acknowledgments}
We acknowledge the use of AI assistance solely for grammar review and the generation of code necessary for producing plots and figures. Any AI-generated content represents a paraphrase of original material authored by the researchers, aimed at improving the readability of the text.

\bibliography{custom}
\appendix

\section{Experimental Setups}
\label{appendix:experimental_setups}

\subsection{Models}
We employed various language models for different components of our pipeline:
\begin{itemize}
\item \textbf{Llama 3 8B \& Gemma 27B \& 12B}: Used for the demographic surveying and ATP question answering components. Gemma 27B is also utilized for persona generation.
\item \textbf{Phi-4-mini-instruct 4B}: Utilized for both persona generation from social network history and response parsing in the demographic surveying phase.
\item \textbf{Gemini 2.5 Flash \& Claude 3.7 Sonnet \& GPT-4.1}: Used for the inconsistency detection pipeline, accessed through the available APIs on Openrouter.ai platform.
\item \textbf{ModernBERT-base \& Qwen3-Embedding-0.6B}: Used for the link prediction and distribution analysis respectively.
\end{itemize}

\begin{table}[ht]
\centering
\small
\setlength{\tabcolsep}{6pt}
\renewcommand{\arraystretch}{1.2}
\begin{tabular}{l r}
\toprule
\textbf{Model} & \textbf{Human Agreement (\%)} \\
\midrule
Gemini 2.5 Flash & 69 \\
Claude 3.7 Sonnet & 61 \\
GPT-4.1 & 56 \\
\bottomrule
\end{tabular}
\caption{Human agreement rates across different language models.}
\label{tab:human_annotator_aggreement}
\end{table}

\subsection{Hardware and Deployment}
All models, except Gemma-27B, were served on two RTX 6000 GPUs and one RTX 8000 GPU. The RTX 6000 machines each have 24 GB of VRAM and 128 GB of system RAM, while the RTX 8000 machine has 48 GB of VRAM and 256 GB of system RAM. Gemma-27B was served on Vertex AI using two H100 GPUs with 160 GB of VRAM.

\subsection{Software}
LLaMA 3 8B, Gemma 27B and 12B, and Phi-4 Mini Instruct were all served using vLLM with Python 3.10. The CUDA version used across all GPUs was 12.4. For the embedding model, the Sentence-Transformers and Transformers libraries were used.

\subsection{Hyperparameters}
\begin{itemize}
\item For demographic surveying and ATP question answering, we used the default hyperparameters specified in the original Anthology paper.
\item For backstory generation, we set the temperature to 0.1 and limited the maximum number of generated tokens to 1500.
\item We used 42 as the random seed for case studies in both model initialization and dataset splitting.
\item To replicate Anthology, we adopted all hyperparameters reported in their paper and GitHub repository.
\end{itemize}

\section{Dataset Generation}
\label{appendix:dataset_generation}

\begin{figure*}[t]
\centering
\includegraphics[width=1\textwidth]{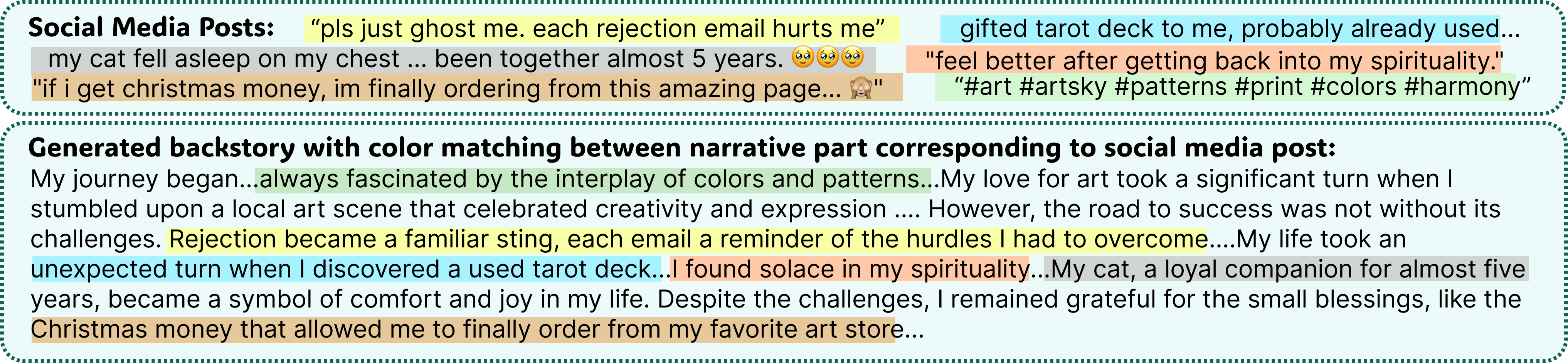}
\caption{Illustrative example from \textsc{Synthia} showing a persona and its grounding social media posts. Highlights demonstrate how different spans of the persona relate to their respective source posts.}
\label{fig:grounding}
\end{figure*}

We clean each of users and corresponding posts by de-duplication, removing posts with unusual dates (such as 1/1/1), and removing non-English posts. We used ``langdetect`` library to label each post with a language.

\begin{figure}[t]
\centering
\includegraphics[width=1\columnwidth]{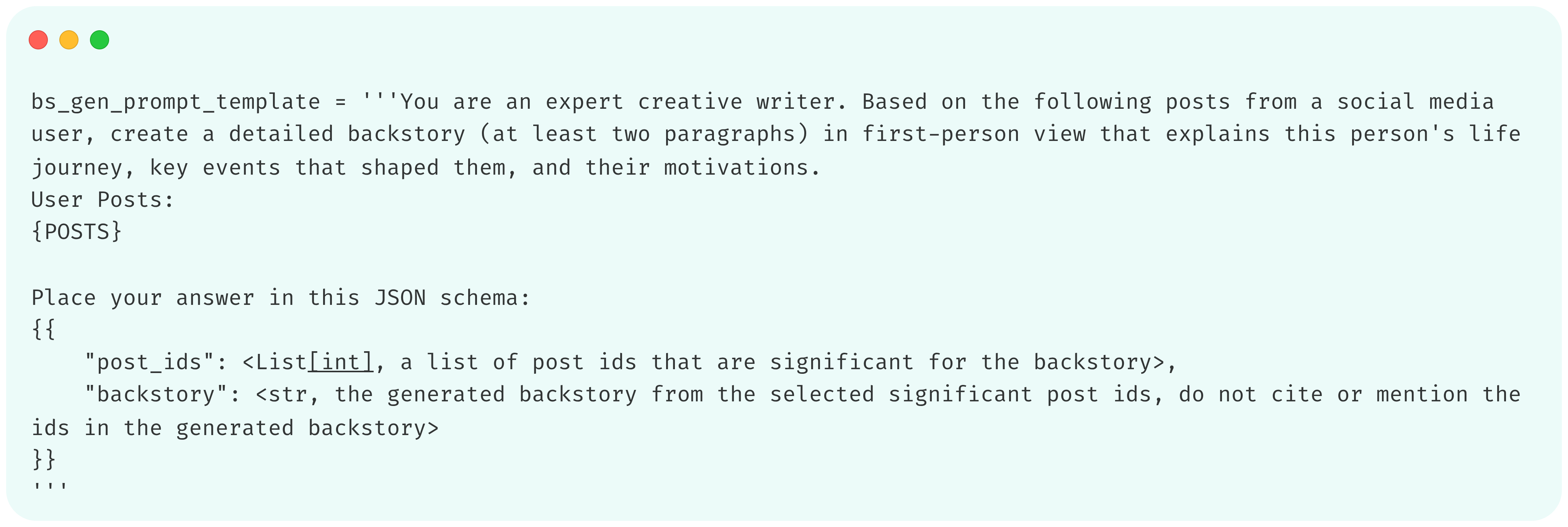}
\caption{Prompt template for backstory generator model.}
\label{fig:pt_gen}
\end{figure}

Unless otherwise specified, \Syn{Gemma} personas are generated with Gemma-3-27B and \Syn{Phi} personas with Phi-4-mini-instruct, with the prompt illustrated in Figure~\ref{fig:pt_gen}.
A sample of personas with grounding data is presented in Figure~\ref{fig:grounding}.

\begin{table}[ht]
\centering
\small
\setlength{\tabcolsep}{4pt}
\renewcommand{\arraystretch}{1.2}
\begin{tabularx}{\columnwidth}{X r r r}
\toprule
\multicolumn{4}{c}{\textbf{Social Media Data Statistics}} \\
\midrule
Total posts & \multicolumn{3}{r}{667,842} \\
Mean posts per user & \multicolumn{3}{r}{234.50 $\pm$ 262.88} \\
Mean words per post & \multicolumn{3}{r}{15.17 $\pm$ 13.97} \\
Max words per post & \multicolumn{3}{r}{100} \\

\midrule
\multicolumn{4}{c}{\textbf{Persona Data Statistics (Words per Persona)}} \\
\midrule
\textbf{Dataset} & \textbf{Min} & \textbf{Max} & \textbf{Mean} \\
\midrule
\Syn{Gemma} & 194 & 559 & 310.83 \\
\Syn{Phi} & 121 & 351 & 257.48 \\
\Ant{LLaMa} & 8 & 1,376 & 376.56 \\
\Ant{Gemma} & 21 & 1,354 & 506.67 \\
\PChat{Human} & 14 & 76 & 32.51 \\

\midrule
\multicolumn{4}{c}{\textbf{Social Network Metrics (Pruned Graph)}} \\
\midrule
\textbf{Metric} & \multicolumn{3}{r}{\textbf{Value}} \\
\midrule
\# nodes ($N$) & \multicolumn{3}{r}{2,053} \\
\# edges ($M$) & \multicolumn{3}{r}{26,467} \\
Network density & \multicolumn{3}{r}{0.0063} \\
Mean clustering coefficient & \multicolumn{3}{r}{0.2695} \\

\midrule
\multicolumn{4}{c}{\textbf{Connectivity (Pruned Graph)}} \\
\midrule
Weakly connected components & \multicolumn{3}{r}{11} \\
Strongly connected components & \multicolumn{3}{r}{502} \\
Largest WCC size & \multicolumn{3}{r}{2,031} \\

\bottomrule
\end{tabularx}
\caption{Synthesized data created by \textsc{Synthia} used in this work. Network statistics are computed on the pruned graph after removing isolated nodes.}
\label{tab:combined_dataset_stats}
\end{table}

\section{Consistency Analysis}
\label{appendix:consistency_analysis}
\begin{figure}[t]
\centering
\includegraphics[width=1\columnwidth]{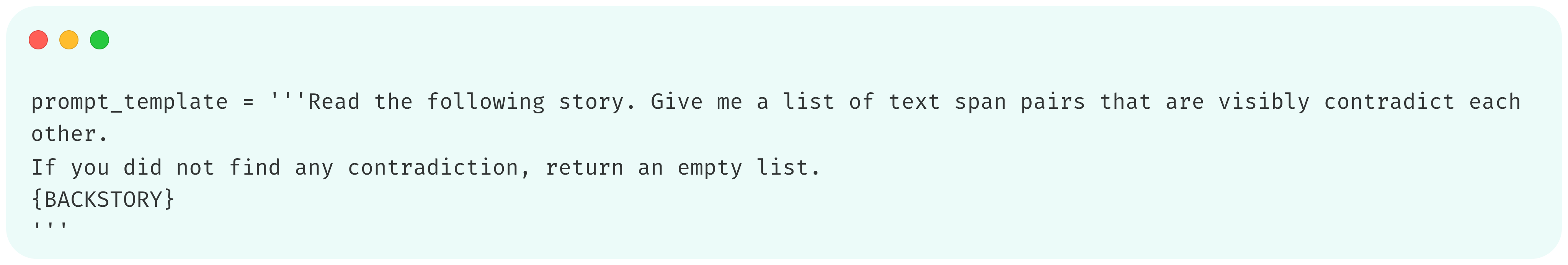}
\caption{Prompt template for inconsistency detector model.}
\label{fig:pt_judge}
\end{figure}

Consistency analysis was conducted with three different API state of the art LLMs and their results were compared to that of humans. ``google/gemini-2.5-flash`` API had the most corelation with human annotators (see Table~\ref{tab:human_annotator_aggreement}), therefore we used this model as our judge to detect contractions across different sets of personas. All the models were accessed through OpenRouter\footnote{http://openrouter.ai/}. Prompt template illustrated in Figure~\ref{fig:pt_judge}. The said model respected the output format for nearly all the cases and the following regex pattern used to parse the outputed JSON from the Judge model:
\begin{center} 
\verb|```json\s*(.*?)\s*```|
\end{center}

\section{Bias Analysis}
\label{appendix:fidelity_gap}

This section details the mathematical formulations used to evaluate the fairness and fidelity of the \textsc{Synthia} personas in Section~\ref{sec:bias}. We introduce two key metrics: the \textit{Fidelity Gap}, which measures accuracy relative to natural human variance, and the \textit{Parity Gap}, which quantifies the disparity in performance across different demographic subgroups.

\subsection{Fidelity Gap}

Standard distance metrics like Earth Mover's Distance (EMD) can be misleading if they do not account for the inherent diversity within a target population. A subgroup with high internal disagreement (e.g., a "purple" state in political polling) is naturally harder to simulate than a homogenous group. To address this, we define the \textbf{Fidelity Gap} as the excess error of the model over the natural internal variation of the human population.

For a demographic category $D$ (e.g., Gender, Age, Race), we first compute the gap for each subgroup $d \in D$, then average across all subgroups as shown in Equation~\ref{eq:fidelity_gap}:

\begin{equation}
\label{eq:fidelity_gap}
\begin{split}
\text{Fidelity Gap}_D = \frac{1}{|D|} \sum_{d \in D} & \Big[ \text{EMD}(P_{\text{human}}^d, P_{\text{LLM}}^d) \\
& - \text{EMD}_{\text{human}}^{\text{internal},d} \Big]
\end{split}
\end{equation}

The critical component here is the \textit{Internal Human EMD} ($\text{EMD}_{\text{human}}^{\text{internal},d}$), which serves as a baseline for the "irreducible" variance within a group. This is calculated via a bootstrapping approach:

\begin{equation}
\label{eq:internal_emd}
\begin{split}
\text{EMD}_{\text{human}}^{\text{internal},d} = \frac{1}{K} \sum_{k=1}^{K} & \frac{1}{|Q|} \sum_{q \in Q} \\
& W_1(P_{\text{human},q}^{d,(k,1)}, P_{\text{human},q}^{d,(k,2)})
\end{split}
\end{equation}

\noindent where:
\begin{itemize}
    \item $D$ is a demographic category (e.g., gender = \{male, female\})
    \item $d \in D$ is a specific subgroup within that category (e.g., male)
    \item $P_{\text{human}}^d$ is the weighted response distribution of humans in subgroup $d$
    \item $P_{\text{LLM}}^d$ is the response distribution of LLM personas assigned to subgroup $d$
    \item $\text{EMD}_{\text{human}}^{\text{internal},d}$ represents the lower-bound proxy for natural human variance within subgroup $d$
    \item $K$ is the number of random splits (we use $K=20$)
    \item $Q$ is the set of all questions of interest
    \item $W_1$ denotes the Wasserstein-1 distance (Earth Mover's Distance)
    \item $P_{\text{human},q}^{d,(k,i)}$ is the weighted response distribution for question $q$ among humans in subgroup $d$ in the $i$-th half of the $k$-th random split
\end{itemize}

To compute $\text{EMD}_{\text{human}}^{\text{internal},d}$ for each subgroup $d$, we randomly split the human population within that subgroup into two equal halves $K$ times. For each split, we calculate the EMD between the two halves across all questions, then average over all splits and questions. The final Fidelity Gap for demographic category $D$ is the average of the individual subgroup gaps.

\subsection{Parity Gap}
\label{appendix:parity_gap}

To ensure that the model does not disproportionately fail for specific minority or marginalized groups, we utilize the \textbf{Parity Gap} ($P_D$). This metric captures the "worst-case" disparity in simulation quality within a demographic category.

Formally, for a given demographic $D$ containing a set of subgroups $S$, let $EMD_s$ be the Earth Mover's Distance between the human and model distributions for subgroup $s \in S$. The Parity Gap is defined as the range between the best-performing and worst-performing subgroups:

\begin{equation}
\label{eq:parity_gap}
 P_D = \max_{s \in S} (EMD_s) - \min_{s \in S} (EMD_s)
\end{equation}

A lower $P_D$ indicates a more equitable model where fidelity is consistent regardless of the specific subgroup (e.g., the model simulates high-income and low-income individuals with comparable accuracy), minimizing representational harms.

\section{Social Network Analysis Details}
\label{appendix:sna_details}

This appendix provides the technical specifications for the graph sampling strategies, baseline construction, and the link prediction methodology used in Section~\ref{sec:cases}.

\subsection{Graph Sampling Strategy}
To capture different topological properties of the social network, we extract two distinct test subgraphs from the primary ground-truth graph $G=(V,E)$.

\paragraph{Global Subgraph ($G_{\text{global}}$).}
Also referred to as $G_{\text{rand}}$, this subgraph is constructed via uniform random node sampling to represent the global network structure. We define a subset of test nodes $V_{\text{global}} \subset V$ by selecting 10\% of the total nodes uniformly at random. The associated test edge set is defined as all directed edges in $E$ originating from these nodes: 
\[
E_{\text{global}} = \{(u, v) \in E \mid u \in V_{\text{global}}\}
\]

\paragraph{Local Subgraph ($G_{\text{local}}$).}
Also referred to as $G_{\text{conn}}$, this subgraph is designed to test the model's performance in dense, high-homophily neighborhoods. We select a subset of nodes $V_{\text{local}} \subset V$, where $|V_{\text{local}}| \approx 0.1|V|$, such that $V_{\text{local}}$ forms a single connected component within the directed graph. This snowball sampling approach ensures that the test cases represent a localized community structure rather than disparate actors.

\subsection{Baseline Specifications}
To isolate the impact of \textsc{Synthia}'s generative approach, we compare against two extractive baselines derived from users' historical activity. Crucially, both baselines are constrained by a length-matching parameter to ensure parity with the synthesized personas. Let $L(P_{\text{syn}})$ be the token length of the synthesized persona for a given user.

\begin{description}
    \item[Random Extractive Baseline ($B_{\text{rand}}$):] For each user, we perform random sampling without replacement from their chronological post history. Posts are aggregated until the total character length $L(B_{\text{rand}})$ approximately matches $L(P_{\text{syn}})$. This baseline captures the user's "average" historical signal without recency bias.
    
    \item[Recency-Based Baseline ($B_{\text{rec}}$):] We select the user's most recent posts in descending chronological order. This "latest-first" aggregation continues until $L(B_{\text{rec}}) \approx L(P_{\text{syn}})$, capturing the user's most contemporary linguistic patterns and temporal interests.
\end{description}

\subsection{Link Prediction Methodology}
We formulate link prediction as a binary classification task to evaluate the information density of the personas. Given two personas $v_i$ and $v_j$, the model learns the conditional probability of an edge existence $(v_i, v_j) \in E$.

We fine-tune a transformer-based encoder (ModernBERT-base) to minimize the cross-entropy loss for the probability:
\begin{multline}
P((v_i, v_j) \in E \mid \mathbf{x}) = \\
\sigma(\mathbf{W} \cdot \text{enc}(\text{concat}(v_i, v_j)) + b)
\end{multline}
where $\text{enc}(\cdot)$ is the transformer encoder output, $\text{concat}(v_i, v_j)$ represents the concatenated text of the two personas, and $\sigma$ is the sigmoid function.

\paragraph{Statistical Evaluation.}
We report metrics (Accuracy, Precision, Recall, F1) alongside 95\% bootstrap confidence intervals calculated over 10,000 iterations. To determine statistical significance between \textsc{Synthia} and the baselines, we utilize McNemar's test with continuity correction ($\alpha = 0.05$), which is appropriate for comparing paired binary classification results.
\section{ATP Details}
\label{appendix:atp_details}
\subsection{Demographic Matching Algorithm}
Demographic matching, proposed by~\cite{2024_virtual_berkeley_emnlp}, is an algorithm that identifies the closest persona/backstory to a human by comparing demographic traits. This algorithm samples a subpopulation from our backstory database that best demographically represents the human population for an ATP survey. The algorithm creates a bipartite graph where each backstory and real human is represented by a vertex, with edges representing their demographic similarity.

Here is a formal description of the algorithm:

Let vertex set $H = \{h_1, h_2, ..., h_n\}$ represent a set of $n$ humans, while vertex set $V = \{v_1, v_2, ..., v_m\}$ represents a set of $m$ backstories. Each human $h_i = (t_{i1}, t_{i2}, ..., t_{ik})$ consists of $k$ demographic traits, and each backstory $v_j=(P(d_{j1}), P(d_{j2}), ..., P(d_{jk}))$ represents a probability distribution of demographic traits. The edge $e_{ij}\in E$ connects human $h_i$ and backstory $v_j$.

The weight of edge $w(e_{ij})$ is defined as the product of likelihoods that the $j$-th backstory's traits correspond to the demographic traits of the $i$-th real human. Formally:

$$w(e_{ij}) = w(h_i, v_j) = \prod_{l=1}^{k} P(d_{jl}=t_{il})$$

The demographic matching can then be defined as the following optimization problem:

$$\pi: [n] \rightarrow [m]$$
$$\pi^* = \arg\max_\pi \sum_{i=1}^{n} w(h_i, v_{\pi(i)})$$

We implement a greedy matching approach, where it is not required to match each backstory to exactly one human (i.e., humans can share backstories).

\subsection{Demographic Traits}
\begin{figure}[t]
\centering
\includegraphics[width=1\columnwidth]{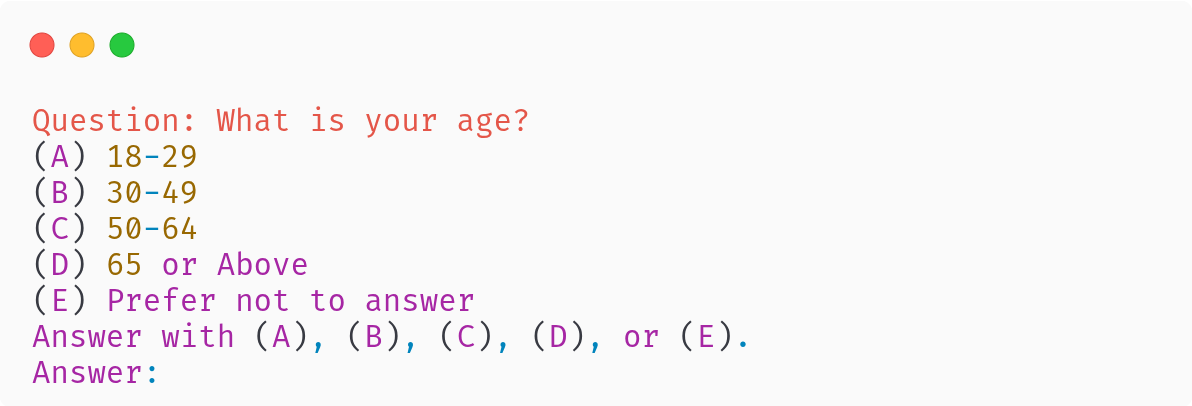}
\caption{Prompt for demographic trait question: age}
\label{fig:demo_q_age}
\end{figure}
\begin{figure}[t]
\centering
\includegraphics[width=1\columnwidth]{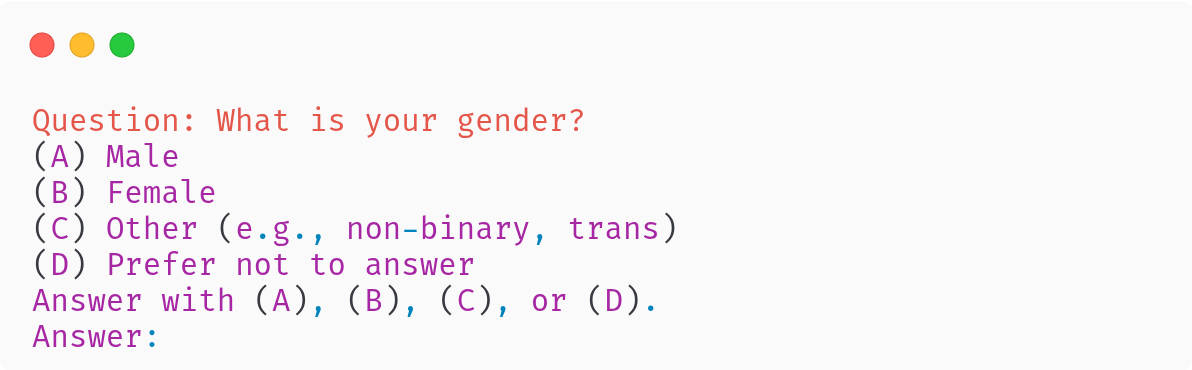}
\caption{Prompt for demographic trait question: gender}
\label{fig:demo_q_gender}
\end{figure}
\begin{figure}[t]
\centering
\includegraphics[width=1\columnwidth]{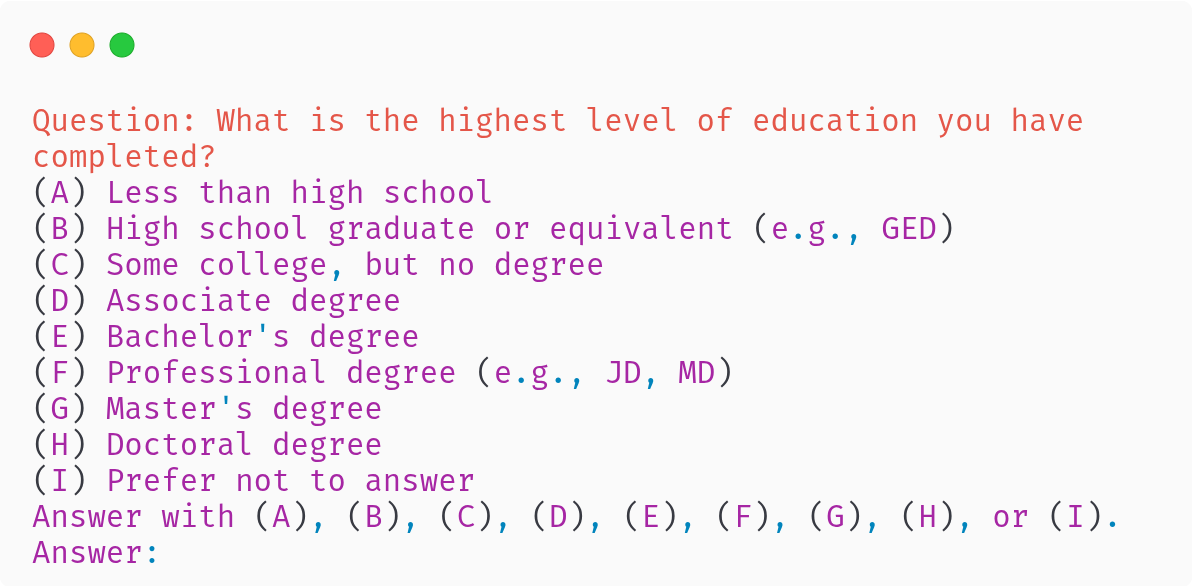}
\caption{Prompt for demographic trait question: education}
\label{fig:demo_q_edu}
\end{figure}
\begin{figure}[t]
\centering
\includegraphics[width=1\columnwidth]{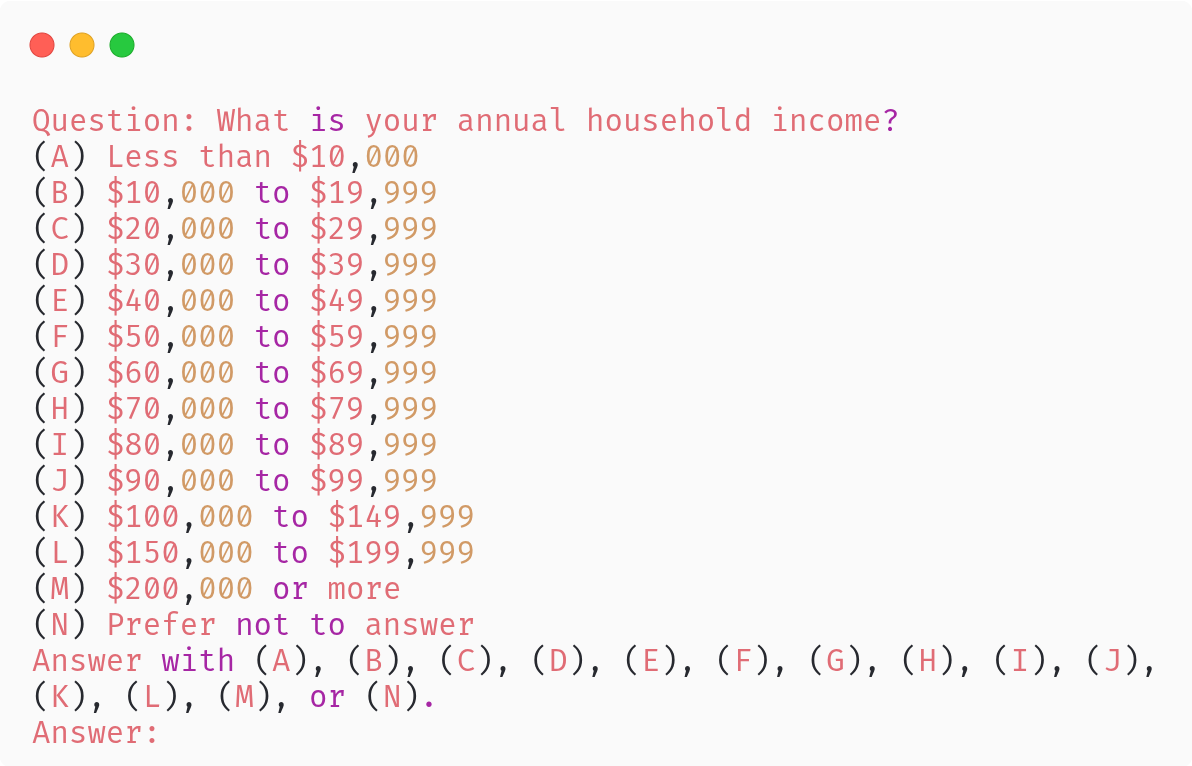}
\caption{Prompt for demographic trait question: income}
\label{fig:demo_q_income}
\end{figure}
\begin{figure}[t]
\centering
\includegraphics[width=1\columnwidth]{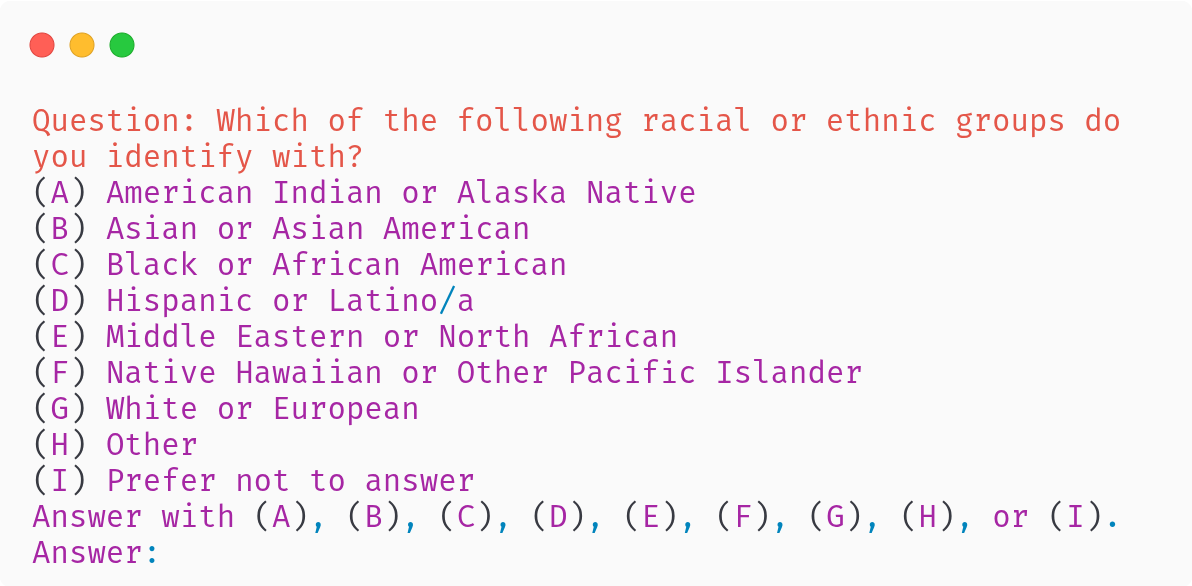}
\caption{Prompt for demographic trait question: race and ethnicity}
\label{fig:demo_q_race}
\end{figure}

In total we have five demographic traits. For each of these a question has been created and asked a non-instruct LLM to answer it 40 times. See Figure~\ref{fig:demo_q_age} for age, Figure~\ref{fig:demo_q_gender} for gender, Figure~\ref{fig:demo_q_edu} for education, Figure~\ref{fig:demo_q_income} for income, and Figure~\ref{fig:demo_q_race} for race and ethnicity.

\subsection{Waves}
Below are the questions of each ATP waves used in this study. The prompt templates are those used in the Anthology \cite{2024_virtual_berkeley_emnlp}.
\label{appendix:wave_qs}
Below is the exact questions and options used for surveying ATP waves in simulations.
\subsubsection{Wave 34}
\begin{itemize}
    \item \textbf{Affordability of GMOs:}
    ``How likely is it that genetically modified foods will lead to more affordably-priced food?''
    \begin{itemize}
        \item Very likely
        \item Fairly likely
        \item Not too likely
        \item Not at all likely
    \end{itemize}

    \item \textbf{Health Problems from GMOs:}
    ``How likely is it that genetically modified foods will lead to health problems for the population as a whole?''
    \begin{itemize}
        \item Very likely
        \item Fairly likely
        \item Not too likely
        \item Not at all likely
    \end{itemize}

    \item \textbf{Environmental Impact of GMOs:}
    ``How likely is it that genetically modified foods will create problems for the environment?''
    \begin{itemize}
        \item Very likely
        \item Fairly likely
        \item Not too likely
        \item Not at all likely
    \end{itemize}

    \item \textbf{Personal Concern (GMOs):}
    ``How much do you, personally, care about the issue of genetically modified foods?''
    \begin{itemize}
        \item A great deal
        \item Some
        \item Not too much
        \item Not at all
    \end{itemize}

    \item \textbf{Organic Consumption:}
    ``How much of the food you eat is organic?''
    \begin{itemize}
        \item Most of it
        \item Some of it
        \item Not too much
        \item None at all
    \end{itemize}

    \item \textbf{Antibiotics and Hormones:}
    ``How much health risk, if any, does eating meat from animals that have been given antibiotics or hormones have for the average person over the course of their lifetime?''
    \begin{itemize}
        \item A great deal of health risk
        \item Some health risk
        \item Not too much health risk
        \item No health risk at all
    \end{itemize}

    \item \textbf{Artificial Coloring:}
    ``How much health risk, if any, does eating food and drinks with artificial coloring have for the average person over the course of their lifetime?''
    \begin{itemize}
        \item A great deal of health risk
        \item Some health risk
        \item Not too much health risk
        \item No health risk at all
    \end{itemize}

    \item \textbf{Artificial Preservatives:}
    ``How much health risk, if any, does eating food and drinks with artificial preservatives have for the average person over the course of their lifetime?''
    \begin{itemize}
        \item A great deal of health risk
        \item Some health risk
        \item Not too much health risk
        \item No health risk at all
    \end{itemize}
\end{itemize}

\subsubsection{Wave 99}
\begin{itemize}
    \item \textbf{AI Knowing Thoughts and Behaviors:}
    ``How excited or concerned would you be if artificial intelligence computer programs could know people's thoughts and behaviors?''
    \begin{itemize}
        \item Very excited
        \item Somewhat excited
        \item Equal excitement and concern
        \item Somewhat concerned
        \item Very concerned
    \end{itemize}

    \item \textbf{AI Performing Household Chores:}
    ``How excited or concerned would you be if artificial intelligence computer programs could perform household chores?''
    \begin{itemize}
        \item Very excited
        \item Somewhat excited
        \item Equal excitement and concern
        \item Somewhat concerned
        \item Very concerned
    \end{itemize}

    \item \textbf{AI Making Important Life Decisions:}
    ``How excited or concerned would you be if artificial intelligence computer programs could make important life decisions for people?''
    \begin{itemize}
        \item Very excited
        \item Somewhat excited
        \item Equal excitement and concern
        \item Somewhat concerned
        \item Very concerned
    \end{itemize}

    \item \textbf{AI Diagnosing Medical Problems:}
    ``How excited or concerned would you be if artificial intelligence computer programs could diagnose medical problems?''
    \begin{itemize}
        \item Very excited
        \item Somewhat excited
        \item Equal excitement and concern
        \item Somewhat concerned
        \item Very concerned
    \end{itemize}

    \item \textbf{AI Performing Repetitive Workplace Tasks:}
    ``How excited or concerned would you be if artificial intelligence computer programs could perform repetitive workplace tasks?''
    \begin{itemize}
        \item Very excited
        \item Somewhat excited
        \item Equal excitement and concern
        \item Somewhat concerned
        \item Very concerned
    \end{itemize}

    \item \textbf{AI Handling Customer Service:}
    ``How excited or concerned would you be if artificial intelligence computer programs could handle customer service calls?''
    \begin{itemize}
        \item Very excited
        \item Somewhat excited
        \item Equal excitement and concern
        \item Somewhat concerned
        \item Very concerned
    \end{itemize}
\end{itemize}

\section{Opinion Alignment Full Results}
\label{appendix:opinion_alignment_res}

\begin{figure*}[t]
    \centering
    \begin{subfigure}[b]{0.48\textwidth}
        \centering
        \includegraphics[width=\linewidth]{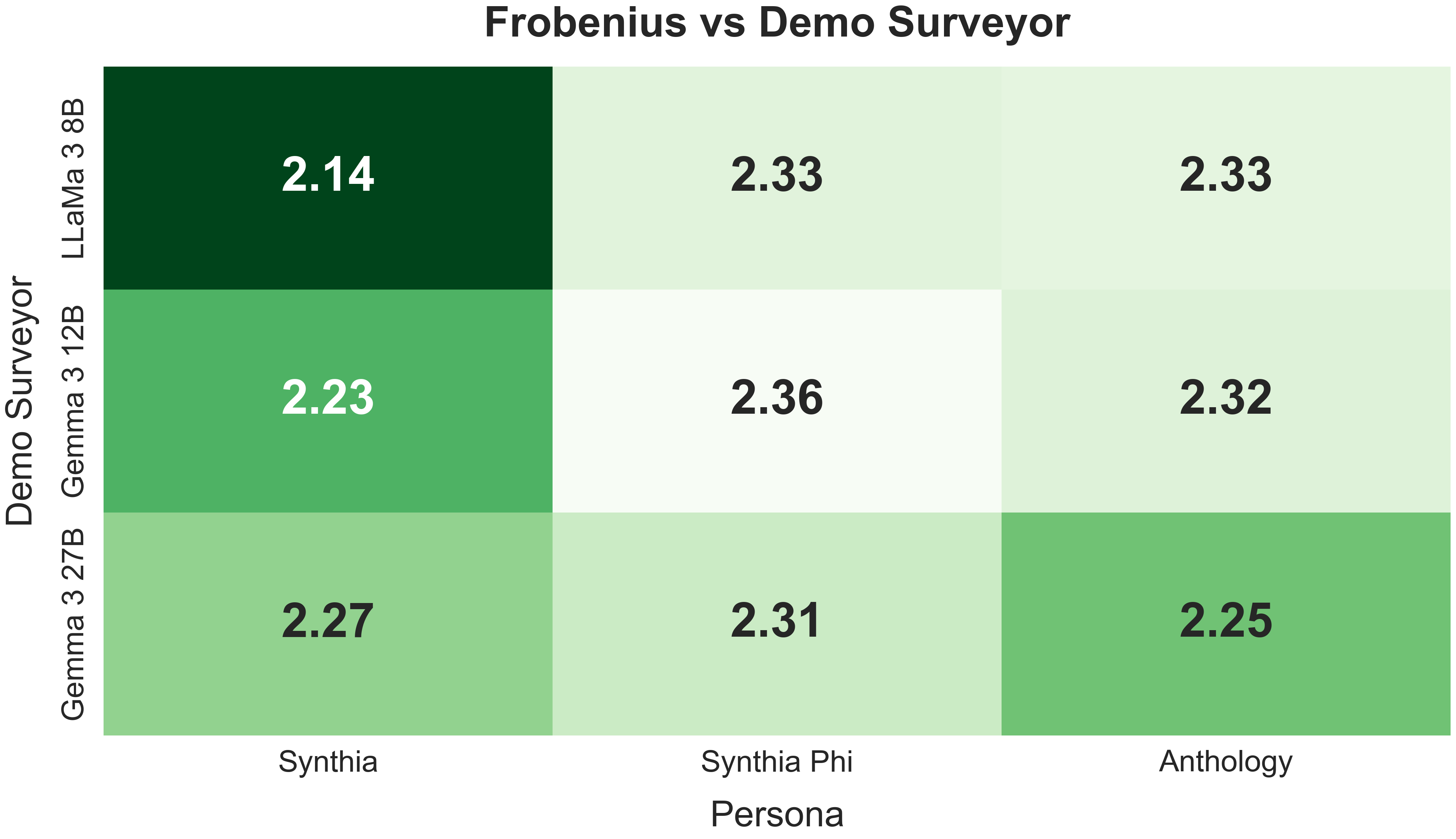}
        \caption{Frobenius Norm by Demographic Surveyor}
        \label{fig:frob_demo}
    \end{subfigure}
    \hfill
    \begin{subfigure}[b]{0.48\textwidth}
        \centering
        \includegraphics[width=\linewidth]{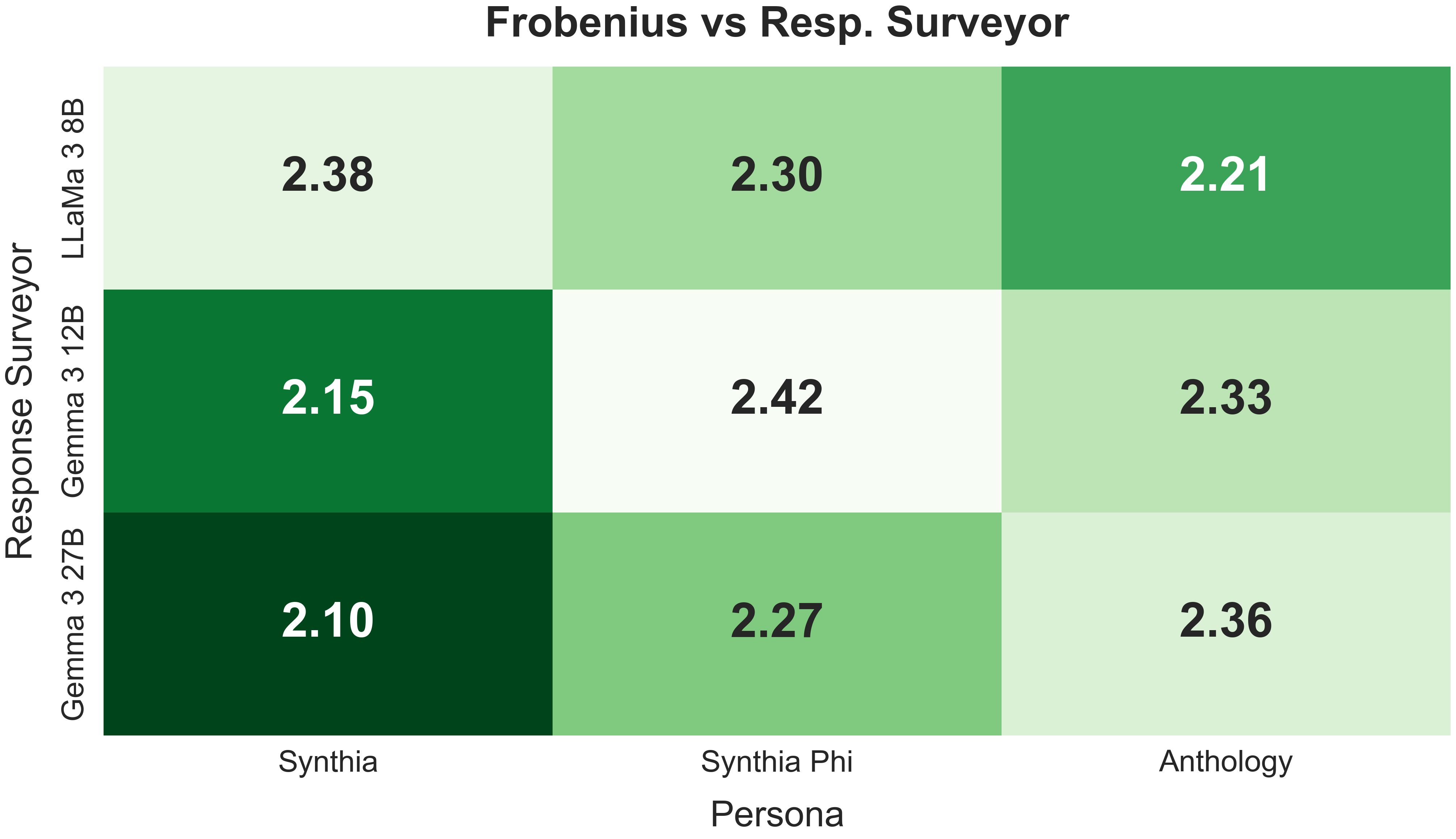}
        \caption{Frobenius Norm by Response Surveyor}
        \label{fig:frob_resp}
    \end{subfigure}
    
    \caption{Frobenius Norm performance heatmaps. Left: Aggregated by Demographic Surveyor. Right: Aggregated by Response Surveyor. Lower values indicate better alignment with the ground-truth correlation matrix.}
    \label{fig:frob_heatmaps}
\end{figure*}

\begin{figure*}[t]
    \centering
    \begin{subfigure}[b]{0.48\textwidth}
        \centering
        \includegraphics[width=\linewidth]{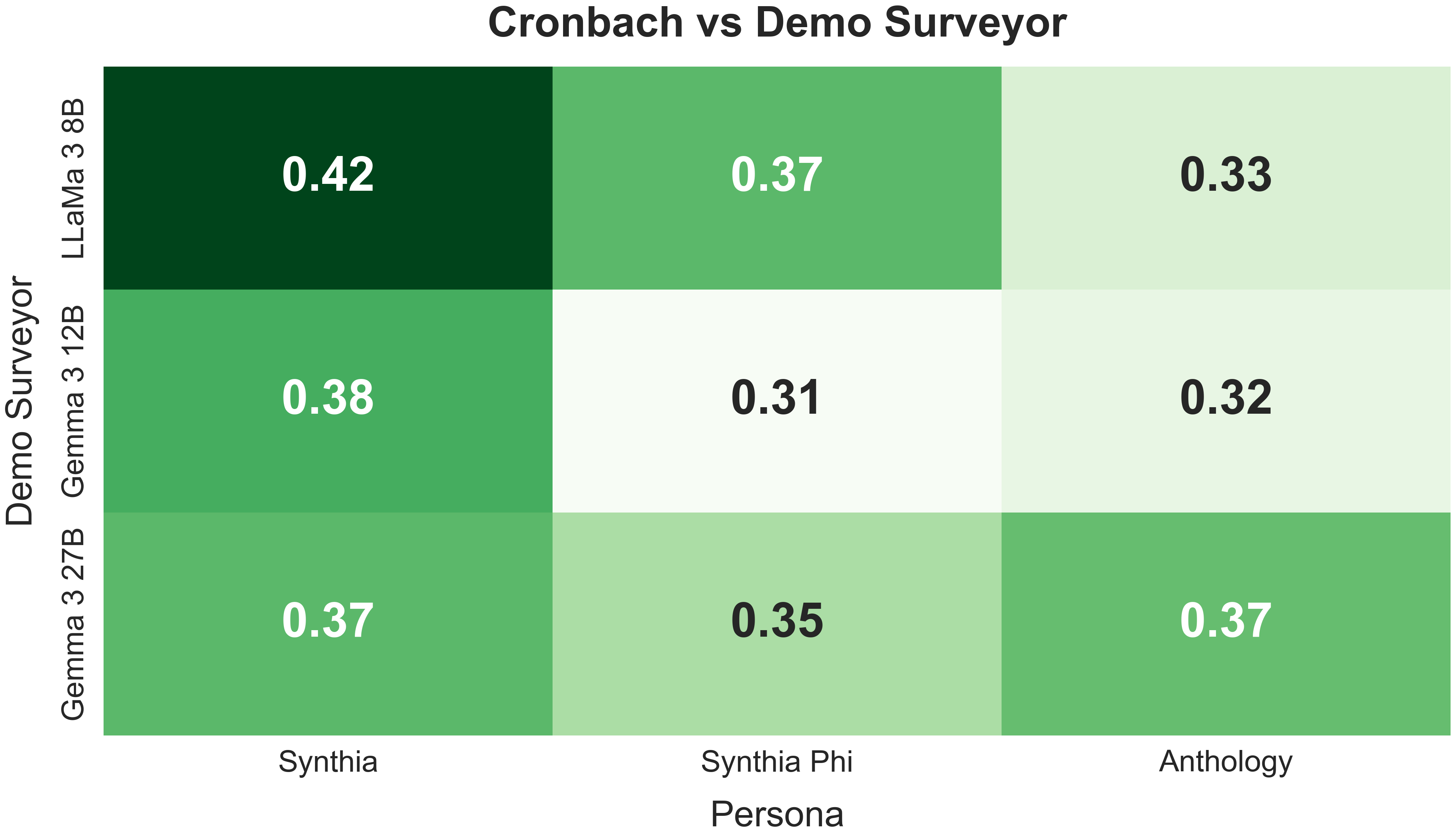}
        \caption{Cronbach's $\alpha$ by Demographic Surveyor}
        \label{fig:cron_demo}
    \end{subfigure}
    \hfill
    \begin{subfigure}[b]{0.48\textwidth}
        \centering
        \includegraphics[width=\linewidth]{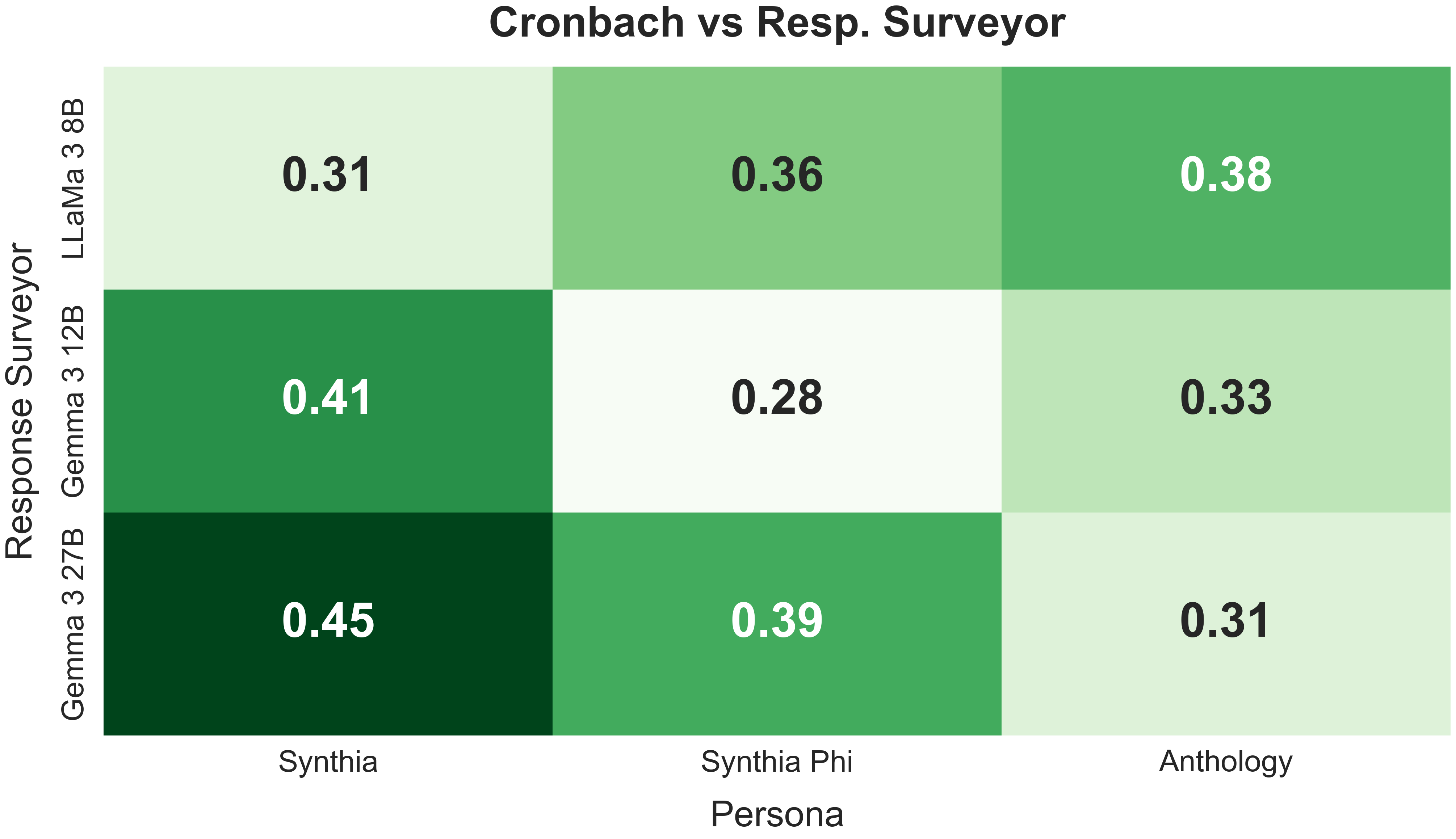}
        \caption{Cronbach's $\alpha$ by Response Surveyor}
        \label{fig:cron_resp}
    \end{subfigure}
    
    \caption{Cronbach's $\alpha$ reliability heatmaps. Left: Aggregated by Demographic Surveyor. Right: Aggregated by Response Surveyor. Higher values indicate greater internal consistency.}
    \label{fig:cron_heatmaps}
\end{figure*}

This section presents the comprehensive quantitative results for our opinion alignment analysis. We report Earth Mover's Distance (EMD), Frobenius norm, and Cronbach's $\alpha$ to evaluate the alignment accuracy and internal consistency of the surveyor models. Tables~\ref{tab:wave_34} and~\ref{tab:wave_99} provide a granular breakdown of interaction dynamics for Wave 34 and Wave 99, respectively, differentiating between demographic and response surveyors across standard backstories. Table~\ref{tab:aggregated} summarizes these findings with an aggregated performance view across all waves. Finally, Table~\ref{tab:gemma12_waves_styled} offers the full results for screening stage. The heatmaps on the relation between Cronbach Alpha and Frobenius Norm metrics with various models as demographic surveyor or response surveyor are given in Figure~\ref{fig:frob_heatmaps} and Figure~\ref{fig:cron_heatmaps}

\begin{figure}[t]
    \centering
    \includegraphics[width=0.8\linewidth]{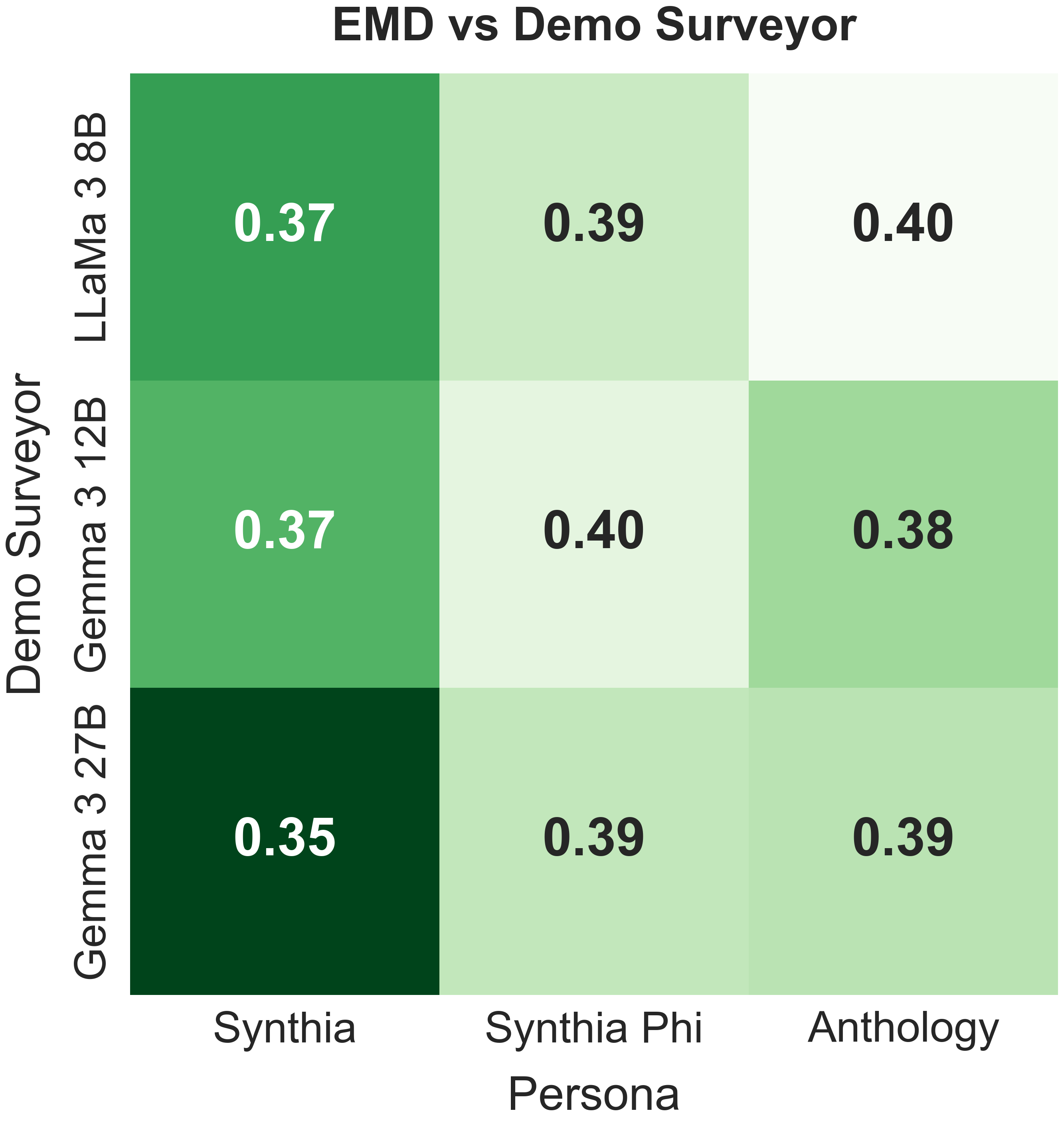}
    \caption{EMD per demo surveyor. Lower is better.}
    \label{fig:emd_demo}
\end{figure}

\begin{figure}[t]
    \centering
    \includegraphics[width=0.8\linewidth]{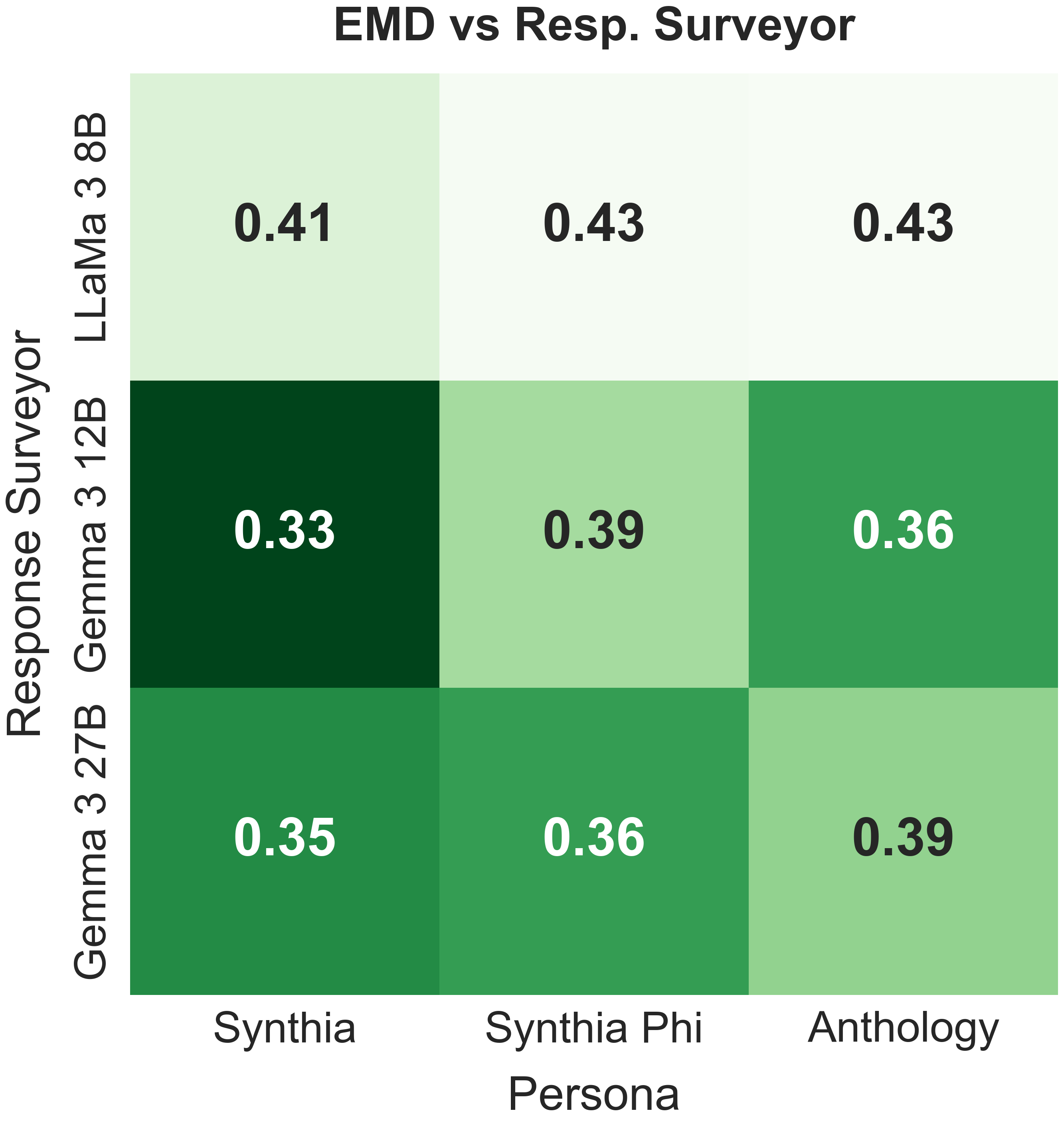}
    \caption{EMD per response surveyor. Lower is better.}
    \label{fig:emd_resp}
\end{figure}

\begin{table}[t]
\centering
\small
\setlength{\tabcolsep}{6pt} 
\caption{Gemma 12 Performance Comparison across Waves 34 and 99}
\label{tab:gemma12_waves_styled}
\begin{tabular}{llccc}
\toprule
\textbf{Wave} & \textbf{Backstory} & \textbf{EMD $\downarrow$} & \textbf{Frob. $\downarrow$} & \textbf{Cron. $\alpha$ $\uparrow$} \\
\midrule

\multirow{5}{*}{34} 
& ant & \textbf{0.341} & \underline{2.407} & \underline{0.348} \\
& ant\_g27 & \underline{0.341} & 2.649 & 0.315 \\
& bsky & 0.362 & \textbf{2.246} & \textbf{0.381} \\
& bsky\_phi & 0.382 & 2.607 & 0.308 \\
& \PChat{Human} & 0.350 & 2.760 & 0.285 \\
\midrule

\rowcolor{gray!10}
& ant & \underline{0.372} & \textbf{2.025} & \textbf{0.412} \\
\rowcolor{gray!10}
& ant\_g27 & 0.486 & 2.412 & 0.197 \\
\rowcolor{gray!10}
& bsky & \textbf{0.338} & 2.210 & 0.337 \\
\rowcolor{gray!10}
& bsky\_phi & 0.449 & 2.393 & 0.167 \\
\rowcolor{gray!10}
\multirow{-5}{*}{99} & \PChat{Human} & 0.578 & \underline{2.048} & \underline{0.380} \\

\bottomrule
\end{tabular}
\end{table}

\begin{table*}[t]
\centering
\small
\setlength{\tabcolsep}{4pt}
\caption{Wave 34: Detailed Interaction Analysis (Ant, Bsky, Bsky-Phi)}
\label{tab:wave_34}
\begin{tabular}{llllccc}
\toprule
\textbf{Wave} & \textbf{Demo Surveyor} & \textbf{Resp. Surveyor} & \textbf{Backstory} & \textbf{EMD $\downarrow$} & \textbf{Frob. $\downarrow$} & \textbf{Cron. $\alpha$ $\uparrow$} \\
\midrule
\multirow{9}{*}{34} & \multirow{9}{*}{gemma\_12} & \multirow{3}{*}{gemma\_12} & ant & \textbf{0.341} & \underline{2.407} & \underline{0.348} \\
& & & bsky & \underline{0.362} & \textbf{2.246} & \textbf{0.381} \\
& & & bsky\_phi & 0.382 & 2.607 & 0.308 \\
\cmidrule(lr){3-7}
& & \multirow{3}{*}{gemma\_27} & \cellcolor{gray!10}ant & \cellcolor{gray!10}\textbf{0.325} & \cellcolor{gray!10}2.474 & \cellcolor{gray!10}0.378 \\
& & & \cellcolor{gray!10}bsky & \cellcolor{gray!10}0.335 & \cellcolor{gray!10}\textbf{2.162} & \cellcolor{gray!10}\textbf{0.482} \\
& & & \cellcolor{gray!10}bsky\_phi & \cellcolor{gray!10}\underline{0.332} & \cellcolor{gray!10}\underline{2.327} & \cellcolor{gray!10}\underline{0.426} \\
\cmidrule(lr){3-7}
& & \multirow{3}{*}{llama8} & ant & \textbf{0.359} & \underline{2.449} & \underline{0.345} \\
& & & bsky & \underline{0.402} & 2.499 & 0.233 \\
& & & bsky\_phi & 0.425 & \textbf{2.425} & \textbf{0.370} \\
\midrule
\rowcolor{gray!10}
& & & ant & \textbf{0.349} & 2.506 & 0.239 \\
\rowcolor{gray!10}
& & & bsky & \underline{0.349} & \textbf{2.229} & \textbf{0.430} \\
\rowcolor{gray!10}
& & \multirow{-3}{*}{gemma\_12} & bsky\_phi & 0.394 & \underline{2.352} & \underline{0.416} \\
\cmidrule(lr){3-7} 
\rowcolor{gray!10}
& & & \cellcolor{white}ant & \cellcolor{white}\underline{0.344} & \cellcolor{white}2.519 & \cellcolor{white}0.332 \\
\rowcolor{gray!10}
& & & \cellcolor{white}bsky & \cellcolor{white}\textbf{0.294} & \cellcolor{white}\textbf{2.357} & \cellcolor{white}\textbf{0.461} \\
\rowcolor{gray!10}
& & \multirow{-3}{*}{\cellcolor{white}gemma\_27} & \cellcolor{white}bsky\_phi & \cellcolor{white}0.347 & \cellcolor{white}\underline{2.426} & \cellcolor{white}\underline{0.386} \\
\cmidrule(lr){3-7} 
\rowcolor{gray!10}
& & & ant & \underline{0.385} & \textbf{2.447} & \textbf{0.379} \\
\rowcolor{gray!10}
& & & bsky & \textbf{0.366} & 2.532 & 0.291 \\
\rowcolor{gray!10}
\multirow{-9}{*}{34} & \multirow{-9}{*}{gemma\_27} & \multirow{-3}{*}{llama8} & bsky\_phi & 0.402 & \underline{2.474} & \underline{0.326} \\
\midrule
\multirow{9}{*}{34} & \multirow{9}{*}{llama8} & \multirow{3}{*}{gemma\_12} & ant & \underline{0.351} & 2.540 & 0.289 \\
& & & bsky & \textbf{0.319} & \textbf{2.096} & \textbf{0.457} \\
& & & bsky\_phi & 0.352 & \underline{2.500} & \underline{0.306} \\
\cmidrule(lr){3-7}
& & \multirow{3}{*}{gemma\_27} & \cellcolor{gray!10}ant & \cellcolor{gray!10}0.379 & \cellcolor{gray!10}2.384 & \cellcolor{gray!10}0.349 \\
& & & \cellcolor{gray!10}bsky & \cellcolor{gray!10}\textbf{0.337} & \cellcolor{gray!10}\textbf{2.043} & \cellcolor{gray!10}\underline{0.454} \\
& & & \cellcolor{gray!10}bsky\_phi & \cellcolor{gray!10}\underline{0.347} & \cellcolor{gray!10}\underline{2.286} & \cellcolor{gray!10}\textbf{0.477} \\
\cmidrule(lr){3-7}
& & \multirow{3}{*}{llama8} & ant & \textbf{0.353} & \textbf{2.380} & \underline{0.361} \\
& & & bsky & \underline{0.385} & 2.577 & 0.333 \\
& & & bsky\_phi & 0.439 & \underline{2.492} & \textbf{0.392} \\
\bottomrule
\end{tabular}
\end{table*}

\begin{table*}[t]
\centering
\small
\setlength{\tabcolsep}{4pt}
\caption{Wave 99: Detailed Interaction Analysis (Ant, Bsky, Bsky-Phi)}
\label{tab:wave_99}
\begin{tabular}{llllccc}
\toprule
\textbf{Wave} & \textbf{Demo Surveyor} & \textbf{Resp. Surveyor} & \textbf{Backstory} & \textbf{EMD $\downarrow$} & \textbf{Frob. $\downarrow$} & \textbf{Cron. $\alpha$ $\uparrow$} \\
\midrule
\multirow{9}{*}{99} & \multirow{9}{*}{gemma\_12} & \multirow{3}{*}{gemma\_12} & ant & \underline{0.372} & \textbf{2.025} & \textbf{0.412} \\
& & & bsky & \textbf{0.338} & \underline{2.210} & \underline{0.337} \\
& & & bsky\_phi & 0.449 & 2.393 & 0.167 \\
\cmidrule(lr){3-7}
& & \multirow{3}{*}{gemma\_27} & \cellcolor{gray!10}ant & \cellcolor{gray!10}0.438 & \cellcolor{gray!10}2.442 & \cellcolor{gray!10}0.136 \\
& & & \cellcolor{gray!10}bsky & \cellcolor{gray!10}\underline{0.404} & \cellcolor{gray!10}\textbf{2.041} & \cellcolor{gray!10}\textbf{0.441} \\
& & & \cellcolor{gray!10}bsky\_phi & \cellcolor{gray!10}\textbf{0.352} & \cellcolor{gray!10}\underline{2.212} & \cellcolor{gray!10}\underline{0.301} \\
\cmidrule(lr){3-7}
& & \multirow{3}{*}{llama8} & ant & 0.467 & \textbf{2.138} & \underline{0.337} \\
& & & bsky & \textbf{0.404} & 2.219 & \textbf{0.386} \\
& & & bsky\_phi & \underline{0.445} & \underline{2.176} & 0.317 \\
\midrule
\rowcolor{gray!10}
& & & ant & \underline{0.390} & \underline{2.158} & \textbf{0.378} \\
\rowcolor{gray!10}
& & & bsky & \textbf{0.286} & \textbf{2.153} & \underline{0.353} \\
\rowcolor{gray!10}
& & \multirow{-3}{*}{gemma\_12} & bsky\_phi & 0.400 & 2.193 & 0.307 \\
\cmidrule(lr){3-7}
\rowcolor{gray!10}
& & & \cellcolor{white}ant & \cellcolor{white}\underline{0.390} & \cellcolor{white}\textbf{2.074} & \cellcolor{white}\textbf{0.424} \\
\rowcolor{gray!10}
& & & \cellcolor{white}bsky & \cellcolor{white}0.395 & \cellcolor{white}2.187 & \cellcolor{white}0.329 \\
\rowcolor{gray!10}
& & \multirow{-3}{*}{\cellcolor{white}gemma\_27} & \cellcolor{white}bsky\_phi & \cellcolor{white}\textbf{0.360} & \cellcolor{white}\underline{2.118} & \cellcolor{white}\underline{0.348} \\
\cmidrule(lr){3-7}
\rowcolor{gray!10}
& & & ant & 0.482 & \textbf{1.731} & \textbf{0.467} \\
\rowcolor{gray!10}
& & & bsky & \textbf{0.427} & \underline{2.123} & \underline{0.357} \\
\rowcolor{gray!10}
\multirow{-9}{*}{99} & \multirow{-9}{*}{gemma\_27} & \multirow{-3}{*}{llama8} & bsky\_phi & \underline{0.445} & 2.215 & 0.290 \\
\midrule
\multirow{9}{*}{99} & \multirow{9}{*}{llama8} & \multirow{3}{*}{gemma\_12} & ant & \underline{0.367} & \underline{2.305} & \underline{0.326} \\
& & & bsky & \textbf{0.312} & \textbf{1.984} & \textbf{0.472} \\
& & & bsky\_phi & 0.375 & 2.429 & 0.205 \\
\cmidrule(lr){3-7}
& & \multirow{3}{*}{gemma\_27} & \cellcolor{gray!10}ant & \cellcolor{gray!10}0.448 & \cellcolor{gray!10}2.274 & \cellcolor{gray!10}0.249 \\
& & & \cellcolor{gray!10}bsky & \cellcolor{gray!10}\textbf{0.363} & \cellcolor{gray!10}\textbf{1.826} & \cellcolor{gray!10}\textbf{0.538} \\
& & & \cellcolor{gray!10}bsky\_phi & \cellcolor{gray!10}\underline{0.438} & \cellcolor{gray!10}\underline{2.231} & \cellcolor{gray!10}\underline{0.385} \\
\cmidrule(lr){3-7}
& & \multirow{3}{*}{llama8} & ant & 0.528 & \underline{2.096} & \underline{0.413} \\
& & & bsky & \underline{0.489} & 2.306 & 0.256 \\
& & & bsky\_phi & \textbf{0.406} & \textbf{1.918} & \textbf{0.482} \\
\bottomrule
\end{tabular}
\end{table*}

\begin{table*}[t]
\centering
\small
\setlength{\tabcolsep}{4pt}
\caption{Aggregated Performance by Demographic and Response Surveyor Models}
\label{tab:aggregated}
\begin{tabular}{llllccc}
\toprule
\textbf{Wave} & \textbf{Demo Surveyor} & \textbf{Resp. Surveyor} & \textbf{Backstory} & \textbf{EMD $\downarrow$} & \textbf{Frob. $\downarrow$} & \textbf{Cron. $\alpha$ $\uparrow$} \\
\midrule
\multirow{9}{*}{All} & \multirow{9}{*}{gemma\_12} & \multirow{3}{*}{gemma\_12} & ant & \underline{0.36} & \textbf{2.22} & \textbf{0.38} \\
& & & bsky & \textbf{0.35} & \underline{2.23} & \underline{0.36} \\
& & & bsky\_phi & 0.42 & 2.50 & 0.24 \\
\cmidrule(lr){3-7}
& & \multirow{3}{*}{gemma\_27} & \cellcolor{gray!10}ant & \cellcolor{gray!10}0.38 & \cellcolor{gray!10}2.46 & \cellcolor{gray!10}0.26 \\
& & & \cellcolor{gray!10}bsky & \cellcolor{gray!10}\underline{0.37} & \cellcolor{gray!10}\textbf{2.10} & \cellcolor{gray!10}\textbf{0.46} \\
& & & \cellcolor{gray!10}bsky\_phi & \cellcolor{gray!10}\textbf{0.34} & \cellcolor{gray!10}\underline{2.27} & \cellcolor{gray!10}\underline{0.36} \\
\cmidrule(lr){3-7}
& & \multirow{3}{*}{llama8} & ant & \underline{0.41} & \textbf{2.29} & \underline{0.34} \\
& & & bsky & \textbf{0.40} & 2.36 & 0.31 \\
& & & bsky\_phi & 0.43 & \underline{2.30} & \textbf{0.34} \\
\midrule
\rowcolor{gray!10}
& & & ant & \underline{0.37} & 2.33 & 0.31 \\
\rowcolor{gray!10}
& & & bsky & \textbf{0.32} & \textbf{2.19} & \textbf{0.39} \\
\rowcolor{gray!10}
& & \multirow{-3}{*}{gemma\_12} & bsky\_phi & 0.40 & \underline{2.27} & \underline{0.36} \\
\cmidrule(lr){3-7}
\rowcolor{gray!10}
& & & \cellcolor{white}ant & \cellcolor{white}0.37 & \cellcolor{white}2.30 & \cellcolor{white}\underline{0.38} \\
\rowcolor{gray!10}
& & & \cellcolor{white}bsky & \cellcolor{white}\textbf{0.34} & \cellcolor{white}\textbf{2.27} & \cellcolor{white}\textbf{0.40} \\
\rowcolor{gray!10}
& & \multirow{-3}{*}{\cellcolor{white}gemma\_27} & \cellcolor{white}bsky\_phi & \cellcolor{white}\underline{0.35} & \cellcolor{white}\underline{2.27} & \cellcolor{white}0.37 \\
\cmidrule(lr){3-7}
\rowcolor{gray!10}
& & & ant & 0.43 & \textbf{2.09} & \textbf{0.42} \\
\rowcolor{gray!10}
& & & bsky & \textbf{0.40} & \underline{2.33} & \underline{0.32} \\
\rowcolor{gray!10}
\multirow{-9}{*}{All} & \multirow{-9}{*}{gemma\_27} & \multirow{-3}{*}{llama8} & bsky\_phi & \underline{0.42} & 2.34 & 0.31 \\
\midrule
\multirow{9}{*}{All} & \multirow{9}{*}{llama8} & \multirow{3}{*}{gemma\_12} & ant & \underline{0.36} & \underline{2.42} & \underline{0.31} \\
& & & bsky & \textbf{0.32} & \textbf{2.04} & \textbf{0.47} \\
& & & bsky\_phi & 0.36 & 2.46 & 0.26 \\
\cmidrule(lr){3-7}
& & \multirow{3}{*}{gemma\_27} & \cellcolor{gray!10}ant & \cellcolor{gray!10}0.41 & \cellcolor{gray!10}2.33 & \cellcolor{gray!10}0.30 \\
& & & \cellcolor{gray!10}bsky & \cellcolor{gray!10}\textbf{0.35} & \cellcolor{gray!10}\textbf{1.93} & \cellcolor{gray!10}\textbf{0.50} \\
& & & \cellcolor{gray!10}bsky\_phi & \cellcolor{gray!10}\underline{0.39} & \cellcolor{gray!10}\underline{2.26} & \cellcolor{gray!10}\underline{0.43} \\
\cmidrule(lr){3-7}
& & \multirow{3}{*}{llama8} & ant & 0.44 & \underline{2.24} & \underline{0.39} \\
& & & bsky & \underline{0.44} & 2.44 & 0.29 \\
& & & bsky\_phi & \textbf{0.42} & \textbf{2.21} & \textbf{0.44} \\
\bottomrule
\end{tabular}
\end{table*}

\subsection{Sensitivity to Likert Scale Resolution}
\label{appendix:likert_sensitivity}

To assess whether the relative performance of persona sets depends on the granularity of the original Likert scale, we perform a sensitivity analysis using a coarser 3-point ordinal formulation. For each survey item, we collapse the original 5-point response categories into three bins while preserving the ordinal structure of the question, and recompute EMD, Frob., and Cron.\ $\alpha$ across the same evaluation settings used in the main analysis.

As shown in Table~\ref{tab:likert_sensitivity}, the qualitative ranking of methods remains stable under this coarser scale. In particular, \Syn{Gemma} remains the strongest overall system in five of the six primary wave-level comparisons. We further quantify agreement between the 5-point and 3-point evaluations in Table~\ref{tab:likert_corr}, observing high correspondence across settings. These results indicate that the improvements of \Syn{Gemma} are robust to response-scale resolution and are not driven by fine-grained calibration artifacts.

\begin{table*}[t!]
\centering
\small
\setlength{\tabcolsep}{5pt}
\scalebox{0.95}{
\begin{tabular}{clccc}
\toprule
\textbf{Wave} & \textbf{Exp.} & \textbf{EMD (5-pt / 3-pt)} $\downarrow$ & \textbf{Frob. (5-pt / 3-pt)} $\downarrow$ & \textbf{Cron.\ $\alpha$ (5-pt / 3-pt)} $\uparrow$ \\
\midrule
34 & \Syn{Gemma} & \textbf{0.3506 / 0.1899} & \textbf{2.3012 / 1.8723} & \textbf{0.3917 / 0.3666} \\
34 & \Ant{LLaMa} & 0.3543 / 0.2049 & 2.4578 / 2.0427 & 0.3351 / 0.2957 \\
34 & \Syn{Phi}   & 0.3805 / 0.2110 & 2.4333 / 1.9997 & 0.3785 / 0.3449 \\
\midrule
99 & \Syn{Gemma} & \textbf{0.3824 / 0.2240} & \textbf{2.0842} / 1.8628 & \textbf{0.3881 / 0.3161} \\
99 & \Ant{LLaMa} & 0.4369 / 0.2506 & 2.1485 / \textbf{1.8518} & 0.3443 / 0.3107 \\
99 & \Syn{Phi}   & 0.4079 / 0.2286 & 2.2112 / 1.9212 & 0.3115 / 0.2686 \\
\bottomrule
\end{tabular}
}
\caption{Sensitivity analysis under a coarser 3-point ordinal response scale. Across the six primary wave-level comparisons (three metrics $\times$ two waves), \Syn{Gemma} remains the strongest overall system in five cases, indicating that the main findings are robust to response-scale resolution.}
\label{tab:likert_sensitivity}
\end{table*}

\begin{table}[t!]
\centering
\small
\setlength{\tabcolsep}{8pt}
\begin{tabular}{lcccc}
\toprule
\textbf{Metric} & \textbf{Mean} & \textbf{Median} & \textbf{Min} & \textbf{Max} \\
\midrule
Pearson $r$ & 0.8937 & 0.9136 & 0.7045 & 0.9820 \\
\bottomrule
\end{tabular}
\caption{Correlation between the 5-point and 3-point evaluations across settings. In addition to the high Pearson correlation shown above, we observe strong rank preservation with Spearman $\rho \approx 0.87$.}
\label{tab:likert_corr}
\end{table}

\end{document}